\documentclass[runningheads]{llncs}
\usepackage{graphicx}
\usepackage{amsmath,amssymb} %
\usepackage{color}

\usepackage{epsfig}
\usepackage{cite}
\usepackage{tabularx}
\usepackage{adjustbox}
\usepackage{threeparttable}
\usepackage{multirow}
\usepackage{subfigure}
\usepackage{caption}
\usepackage{booktabs}

\DeclareMathOperator*{\argmin}{argmin}
\usepackage[dvipsnames]{xcolor}
\newcommand{\bx}{\mathbf{x}}

\newcommand{\mL}{\mathcal{L}}
\newcommand{\mLM}{\mathcal{L}_M}
\newcommand{\mLS}{\mathcal{L}_S}
\newcommand{\lambdaM}{\lambda_M}
\newcommand{\lambdaS}{\lambda_S}

\usepackage[breaklinks=true,bookmarks=false,hidelinks=true]{hyperref}

\makeatletter
\newcommand{\printfnsymbol}[1]{%
  \textsuperscript{\@fnsymbol{#1}}%
}
\makeatother

\begin{document}
\pagestyle{headings}
\mainmatter
\def\ECCVSubNumber{4342}  %

\title{Perceive, Predict, and Plan: Safe Motion Planning Through Interpretable Semantic Representations}

\titlerunning{P3: Safe Motion Planning Through Interpretable Semantic Representations}

\authorrunning{A. Sadat, S. Casas, M. Ren, X. Wu, P. Dhawan, R. Urtasun}
\author{Abbas Sadat\thanks{Denotes equal contribution}$^{1}$, Sergio Casas\printfnsymbol{1}$^{1, 2}$, \\ Mengye Ren$^{1, 2}$, Xinyu Wu$^{1}$, Pranaab Dhawan$^1$, Raquel Urtasun$^{1, 2}$%
\institute{Uber ATG$^1$, University of Toronto$^2$\\
\{asadat, sergio.casas, mren3, xinyuw, pdhawan, urtasun\}@uber.com}
}
\maketitle

\begin{abstract}
  In this paper we propose a novel end-to-end learnable  network that performs joint perception, prediction and motion planning
for self-driving vehicles and produces interpretable intermediate representations. Unlike existing  neural motion planners, our motion planning costs  are  consistent with our perception and prediction estimates. This is achieved by a novel differentiable 
semantic occupancy representation that is explicitly used as cost by the motion planning process.
Our network is learned end-to-end from human demonstrations.
The experiments in a large-scale manual-driving dataset and closed-loop simulation show that the proposed model significantly outperforms state-of-the-art planners
in imitating the human behaviors while producing much safer trajectories.

\end{abstract}
\section{Introduction}

The goal of an autonomy system is to take the output of the sensors, a map, and a  high-level route, and produce a safe and comfortable  ride. 
Meanwhile, producing interpretable  intermediate representations that can explain why the vehicle performed a certain maneuver is very important in safety critical applications such as self-driving, particularly if a bad event was to happen.  
Traditional autonomy stacks produce interpretable representations through the perception and prediction modules in the form of bounding boxes as well as distributions over their future  motion \cite{liang2019mmf,luo2018faf, casas2018intentnet,chai2019multipath,tang2019multiple,casas2019spatially}. 
However, the perception module involves thresholding 
detection confidence scores and running Non-Maximum Supression (NMS) to trade off the precision and recall of the object detector, which cause information loss that could result in  unsafe situations, e.g.,   if a solid object is below the threshold. To handle this, software stacks in industry  rely on a secondary fail safe system that tries to catch all mistakes from perception. This system is however trained separately and it is not easy to decide which system to trust.

First attempts to perform end-to-end neural motion planning did not produce interpretable representations \cite{pomerleau1989alvinn},  and instead focused on producing accurate control outputs that mimic how humans drive \cite{chen2019learning}. %
Recent approaches \cite{hou2019learning,zeng2019neural, rhinehart2018deep}, have tried to incorporate interpretability. The neural motion planner of \cite{zeng2019neural} shared feature representations between perception, prediction and motion planning. However it can produce inconsistent estimates between the modules, as it is framed  as a multi-task learning problem with separate headers between the tasks. As a consequence, the motion planner might ignore detections or motion forecasts, resulting in unsafe behaviors. 

In this paper we take a different approach, and exploit a novel  semantic layer as our intermediate interpretable representation. 
Our approach is designed with safety in mind, and thus does not rely on detection and/or   thresholded activations. Instead, we propose a flexible yet efficient representation that can capture different shapes (not just rectangular objects) and  can handle low-confidence objects. 
In particular, we generate a set of probabilistic semantic occupancy layers over space and time, capturing locations of objects of different classes (i.e., vehicles, bicyclists, and pedestrians) as well as  potentially occluded ones. 
Our motion planner can then use this intermediate representation to penalize maneuvers that intersect regions with higher occupancy probability.
Importantly, our interpretable representation is differentiable, enabling end-to-end learning of the full autonomy system (i.e., from raw sensor data to planned trajectory).
Additionally, as opposed to other neural motion planners \cite{zeng2019neural}, our approach can utilize the intended high-level route not only to plan a trajectory that achieves the goal, but also to further differentiate semantically between on-coming or conflicting traffic. This allows the motion planner to potentially learn the risk with respect to a particular semantic class (e.g.,~moving close to an oncoming vehicle compared to a parked vehicle).%

We demonstrate the effectiveness of our approach on a large-scale dataset that consists of smooth manual-driving in challenging urban scenarios. Furthermore, we use a state-of-the-art sensor simulation to perform closed-loop evaluations of driving behavior produced by our proposed model.
We show that our method is capable of imitating human trajectories more closely than existing approaches while yielding much lower collision rate.

\section{Related Work}

\medskip\noindent\textbf{End-to-end self-driving:}
There is a vast literature on end-to-end approaches to tackle self-driving.
\cite{pomerleau1989alvinn} pioneered this field using a single neural network to directly output
driving control command. More recently, with the success of deep learning, direct control based
methods have advanced with deeper network architectures, more complex sensor inputs, and scalable
learning methods~\cite{bojarski2016end,kendall2018learning,codevilla2018end,muller2018driving}. 
Although directly outputting driving command is a general solution, it may have  stability
and robustness issues \cite{codevilla2019exploring}. Another line of work first outputs the cost map of future trajectories, and then a
trajectory is recovered by looking for local minima on the cost map. The cost map may be parameterized as a simple linear combination of hand crafted costs \cite{sadat2019joint,fan2018auto}, or can be defined in a
general non-parametric form \cite{zeng2019neural}. More recently, cost map based approaches have been
shown to adapt better to more challenging environments. \cite{ma2019navigatewithoutlocalize} proposes to output a 
navigation cost map without localization under a weakly supervised learning environment.
\cite{banzhaf2019nonholonomic} has exploited CNNs to facilitate better sampling in complex driving environments.
\cite{gupta2017cognitive,parisotto2017neural,khan2017memory} explore ways to perceive and map the environment in an end-to-end framework with planning, but do not predict how the world might unroll in the future.
In contrast, our planner relies on interpretable cost terms that use the predicted semantic occupancy maps and hence maintains interpretability and differentiability. 

\medskip\noindent\textbf{Imitation learning and inverse reinforcement learning:}
Our proposed learning algorithm is an instantiation of max-margin planning \cite{maxmarginplan}, which is closely related to imitation learning and inverse reinforcement learning. 
\textit{Imitation learning} attempts to directly regress the control commands from human demonstrations \cite{pomerleau1989alvinn,bojarski2016end,bansal2018chauffeurnet,codevilla2018end, chen2019learning, zhao2019lates, rhinehart2018deep, hawke2019urban}. As this can be a very difficult regression problem needsing large amounts of training data, \cite{muller2018driving} investigates the possibility of transferring knowledge from a simulated environment.

Instead of regressing driving control commands, \textit{max-margin planning} reasons about the
cost associated with each output trajectory \cite{maxmarginplan,fan2018auto,zeng2019neural,sadat2019joint}. It tries
to make the human driving trajectories the least costly among all possible trajectories, and
penalizes for any violations. It also considers the task loss as in any behavioral differences in
the trajectory representation. %
\textit{Inverse reinforcement learning} (IRL) is similar to max-margin planning, where the best trajectory
is replaced by a distribution over trajectories that is characterized by their energy
~\cite{maxentirl, maxentdeep}. Generative adversarial models have also been explored in the field of IRL and imitation  learning \cite{gail}, so that the model learns to generate trajectories that look similar to human demonstrations judged by a classifier network. 

\medskip\noindent\textbf{Multi-task learning:}
Our end-to-end framework adopts multi-task learning, where we train the model on a joint objective of object detection, occupancy forecasting, and motion planning. Multi-task learning has been shown to help  extract more useful information from training data by exploiting task relatedness. \cite{feichtenhofer2017detect,frossard2019deepsignals} showed that detection and tracking can be trained together, and \cite{luo2018faf} applies a joint detector and trajectory predictor into a single model in the context of self-driving. This was further extended by \cite{casas2018intentnet} to also predict the high-level intention of actors. 
More recently, \cite{zeng2019neural} further included a cost map based motion planner in the joint model. These works show that joint learning on a multi-task objective helps individual tasks due to better data utilization and shared features, while saving computation.

\medskip\noindent\textbf{Perception and Motion Prediction:}
The majority of previous works have adopted bounding-box detection and trajectory prediction to reason about the future state of a driving scene \cite{luo2018faf, casas2018intentnet, tang2019multiple, casas2019spatially, phan2019covernet, Liang_2020_CVPR, chai2019multipath, zhao2019multi, casas2020importance, li2020end, casas2020implicit}. As there are multiple possible futures, these methods either generate a fixed number of modes with probabilities and/or draw samples to characterize the trajectory distribution. 
In robotics, occupancy grids have been a popular representation of free space. In \cite{elfes1989using}, a framework is proposed to estimate occupancy probability of each grid-cell independently using range sensor data. 
This approach is later extended in \cite{thrun2003learning} to model dependencies among neighboring cells.
\cite{hoermann2018dynamic} performs dynamic occupancy grid prediction at the scene-level, but it does not predict how the scene might evolve in the future.
\cite{jain2019discrete} proposes a discrete residual flow to predict the distribution over a pedestrian's future position in the form of occupancy maps. 
Similarly, in \cite{ridel2020scene} agent-centric occupancy grids are predicted from past trajectories using ConvLSTMs, and multiple trajectories are then sampled to form possible futures.
\cite{liang2020garden} further improves this output parameterization by predicting a continuous offset to mitigate discretization errors, and proposes a procedure to extract trajectory samples.
Different from these methods, our proposed semantic perception and future prediction is instance-free and directly produces occupancy
layers for the entire scene, rather than for each actor instance, which makes it efficient.
Moreover, since no thresholding is employed on the detection scores, our model allows passing low probability objects to the motion planner and hence improving safety of self-driving.

\label{sec:model}
\begin{figure*}[t]
        \centering
        \includegraphics[width=\textwidth]{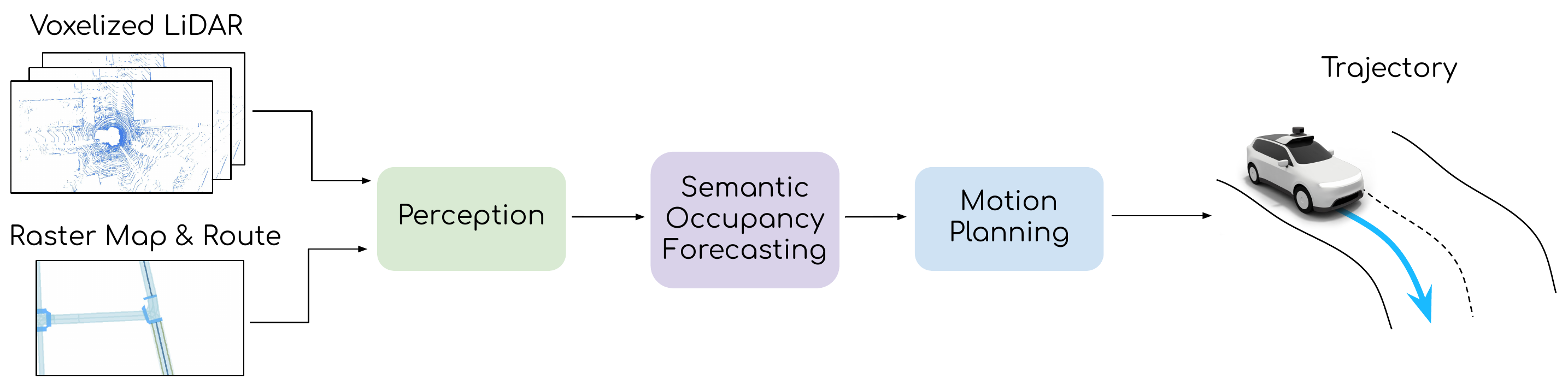}
        \caption{\textbf{Overview} of our proposed end-to-end learnable autonomy system that takes raw sensor data, an HD map and a high level route as input and produces safe maneuvers for the self-driving vehicle via our novel semantic interpretable intermediate representations.}
        \label{fig:overview}
\end{figure*}

\section{End-to-End Interpretable Neural Motion Planner}
In this paper we propose an end-to-end approach to self-driving. 
Importantly, our model produces  intermediate representations that are  designed for safe planning and decision-making, together with interpretability. 
Towards this goal, we exploit the map, the intended route (high level plan to go from point A to point B), and the raw LiDAR point-cloud to generate an intermediate semantic occupancy representation over space and time (i.e., present and future time steps). These interpretable occupancy layers  inform the motion planner about potential objects, including those with low probability,  allowing perception of objects of arbitrary shape, rather than just bounding boxes. 
This is in contrast to  existing approaches \cite{zeng2019neural,casas2018intentnet,bansal2018chauffeurnet} that rely on object detectors that threshold activations and produce  objects with only bounding box shapes. Note that thresholding activations  is very problematic for safety, as if an object is below the threshold it will not be detected, potentially resulting in a collision. 

Our semantic activations are very interpretable. In particular, we generate occupancy layers for each class of vehicles, bicyclists, and pedestrians, as well as occlusion layers which predict occluded objects. 
Furthermore, using the planned route of the self-driving vehicle (SDV), we can semantically differentiate vehicles by their interaction with our intended route (e.g., oncoming traffic vs.~crossing). This not only adds to the interpretability of the perception outputs, but can potentially help the planner learn different subcosts for each category (e.g., different safety buffers for parked vehicles vs oncoming traffic).
We refer the reader to Fig.~\ref{fig:primer} for an exhaustive list of the classes that we predict in the different layers of our occupancy maps.

Our sample-based learnable motion planner then takes these occupancy predictions and evaluates the associated risk of different maneuvers to find a safe and comfortable trajectory for the SDV. This is accomplished through an interpretable  cost function used to cost motion-plan samples, which can efficiently exploit the occupancy information. 
Importantly, our proposed autonomy model is trained end-to-end to imitate human driving while avoiding collisions and traffic infractions. %
Fig. \ref{fig:overview} shows an overview of our proposed approach.

\subsection{Perceiving and Forecasting Semantic Occupancies}

\begin{figure*}[t]
    \centering
    \includegraphics[width=1.0\textwidth]{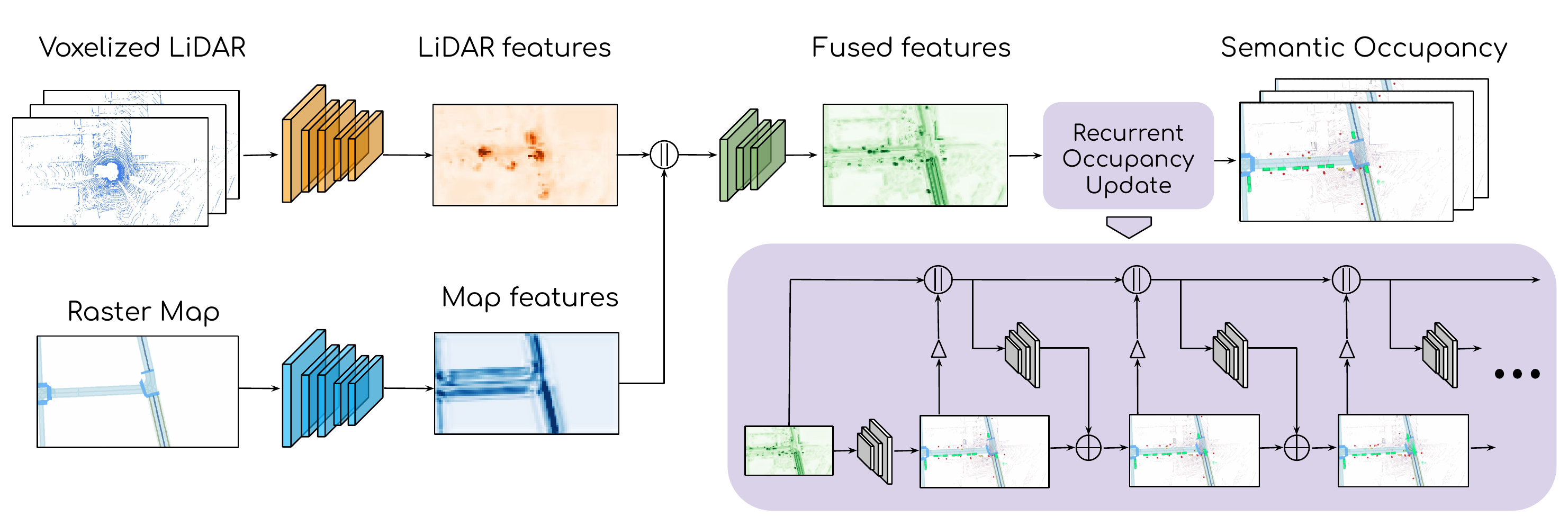}
    \caption{\textbf{Inference diagram} of our proposed perception and recurrent occupancy forecasting model. $\parallel$ symbolizes concatenation along the feature dimension, $\oplus$ element-wise sum and $\bigtriangleup$ bilinear interpolation used to downscale the occupancy.}
    \label{fig:diagram}
\end{figure*}

Our model exploits LiDAR point clouds and HD maps to predict
marginal distributions of semantic occupancy over time, as shown in
Fig. \ref{fig:diagram}. 
These are spatio-temporal, probabilistic, and
instance-free representations of the present and future that capture whether a spatial region is occupied by any dynamic agent belonging to a semantic group at discrete time steps.
Note that this representation naturally captures multi-modality in the future behavior of actors by placing probability mass on different spatial regions at future time steps, which is important as the future might unroll in very different ways (e.g., vehicle in front of the SDV brakes/accelerates, a pedestrian jaywalks/stays in the sidewalk).

\begin{figure*}[t]
    \centering
    \includegraphics[width=1.0\textwidth]{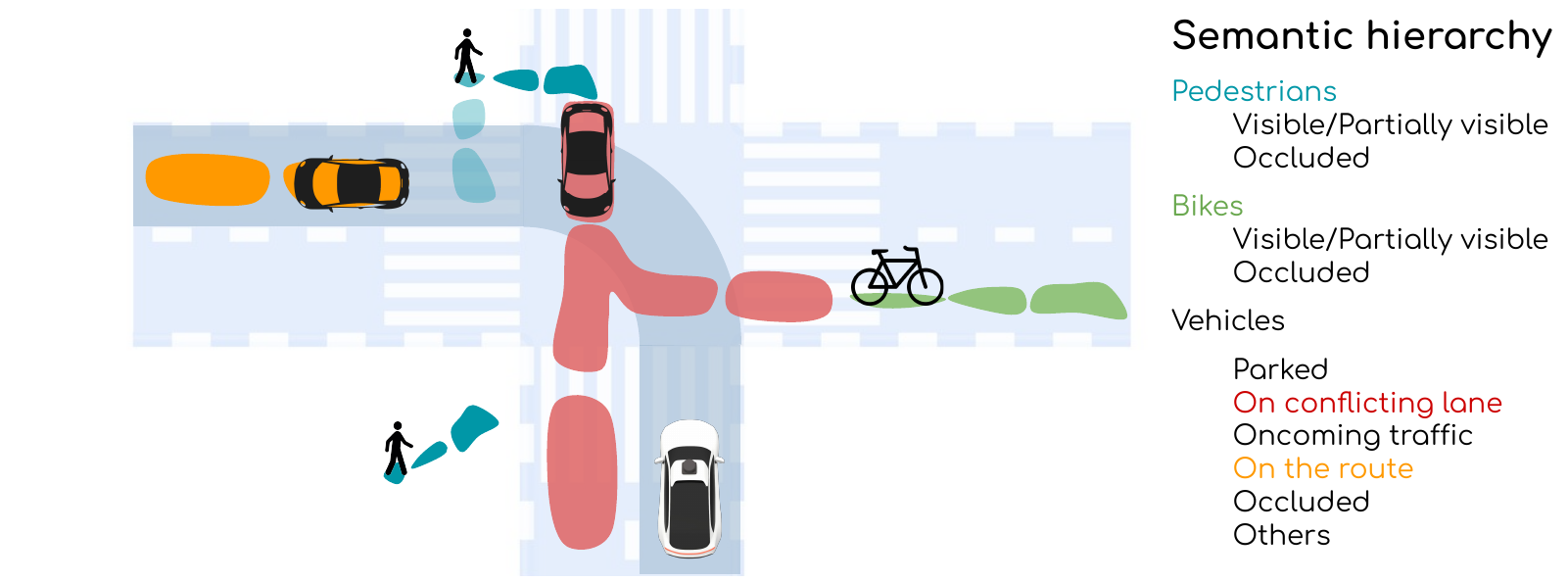}
    \caption{\textbf{Semantic classes} in our occupancy forecasting. Colors match between drawing and hierarchy. Shadowed area corresponds to the SDV route. Black vehicle, pedestrian and bike icons represent the agents' true current location.}
    \label{fig:primer}
\end{figure*}

\medskip\noindent\textbf{Input Representation:}
We use several consecutive LiDAR sweeps as well as HD maps (including  lane graphs) as input to our model as they bring complementary information. %
Following~\cite{zeng2019neural}, we voxelize $T_p$=10 past LiDAR point
clouds in bird's eye view (BEV) with a resolution of $a$=0.2 meters/voxel. Our region of interest is $W$=140m long (70m front and behind of the SDV), $H$=80m wide (40 to each side of the SDV), and $Z$=5m tall; obtaining a 3D tensor of size $(\frac{H}{a}, \frac{W}{a}, \frac{Z}{a}, \cdot T_p)$. As proposed in \cite{casas2018intentnet}, we concatenate height and time along the channel dimension to avoid using 3D convolutions or a recurrent model, thus saving memory and computation. 
Leveraging map information is very important to have a  safe motion planner as we
need to drive according to traffic rules such as stop signs, traffic lights
and lane markers. Maps are also very relevant for perception and motion
forecasting since they provide a strong prior on
the presence as well as the future motion of traffic participants (e.g., vehicles and bikes normally follow lanes, 
pedestrians usually use sidewalks/crosswalks). 
To exploit HD maps, we   adopt the representation
proposed in \cite{casas2018intentnet} and rasterize different semantic
elements (e.g., roads, lanes, intersections,
crossings) into different binary channels to enable separate reasoning about the distinct elements.
For instance, the state of a traffic light (green, yellow, red) is rasterized in 3 different channels, facilitating  traffic flow reasoning at intersections.
All in all,
we obtain a 3D tensor of size $(\frac{H}{a}, \frac{W}{a}, C)$, with $C$=17 binary channels for the map.

\medskip\noindent\textbf{Backbone Network:}
We combine ideas in \cite{yang2018pixor} and
\cite{casas2018intentnet} to build a multi-resolution, two-stream backbone
network that extracts features from the LiDAR voxelization and map raster.
One stream
processes LiDAR while the other one processes the map.
Each stream is composed of 4 residual blocks with number of layers (2, 2, 3, 6) 
and stride (1, 2, 2, 2) respectively. Thus, the features after each residual block are
$\mathcal{F}_{1x}, \mathcal{F}_{2x}, \mathcal{F}_{4x}, \mathcal{F}_{8x}$, where the subscript indicates the downsampling factor from the input in BEV. The features from the different
blocks are then concatenated at 4x downsampling by max pooling higher resolution ones $\mathcal{F}_{1x}, \mathcal{F}_{2x}$ and interpolating $\mathcal{F}_{8x}$, as proposed by \cite{yang2018pixor}. 
The only difference between the two streams is that the LiDAR one uses more features (32, 64, 128, 256) versus (16, 32, 64, 128) on the map stream. We give more capacity to the LiDAR branch as the input is much higher dimensional than the raster map, and the backbone is responsible for aggregating geometric information from different past LiDAR sweeps to extract good appearance and motion cues.
Finally, the LiDAR and map features are  fused by concatenation along the feature dimension followed by a final residual block of 4
convolutional layers with no downsampling, which outputs a tensor $\mathcal{F}$ with 256 features.

\medskip\noindent\textbf{Semantic Occupancy Forecasting:}
Predicting the future motion of traffic participants is very challenging as actors can perform complex motions and there is a lot of uncertainty due to both partial observability (because of sensor occlusion or noise) as well as the multi-modal nature of the possible outcomes. 
Many existing approaches have modeled the underlying distribution in a parametric way (e.g., Gaussian, mixture of Gaussians) \cite{casas2019spatially,chai2019multipath}. While efficient, this incorporates   strong assumptions, lacks expressivity and  is prone to instabilities during optimization (see \cite{jain2019discrete}). 
Jain et al. \cite{jain2019discrete} use non-parametric occupancy distributions for each instance (i.e., actor) naturally capturing complex  multi-modal distributions. However,  this is a computationally and memory inefficient representation that  scales poorly with the number of actors, which can be hundreds in crowded scenes. In contrast, in this paper we propose a  novel representation, where groups of actors are modeled with a single non-parametric distribution of future semantic occupancy. This removes the need for both detection and tracking and is both efficient and effective as shown in our experiments. %

In particular, these actors are grouped semantically in a hierarchy. 
We consider vehicles, pedestrians and bikes as the root semantic classes $\mathcal{C}$, as shown in Fig.  \ref{fig:primer}. For each root category, we consider mutually-exclusive subclasses which include a negative (not occupied) subclass. 
Note that the root categories are not mutually exclusive as actors that belong to different classes can share the same occupied space (e.g., pedestrian getting in or out of a car). 
We create these subdivisions because we wish to learn different planning costs for each of these subclass occupancies, given that such subcategories have very different semantics for driving. For instance, parked vehicles require a smaller safety buffer than a fast moving vehicle in a lane that conflicts with the SDV route, since they are not likely to move and therefore the uncertainty around them is smaller. 
We do not subdivide the pedestrians and bikes (with riders) by their semantic location (road/sidewalk) or behavior (stationary/moving), as they are vulnerable road users and thus we want to make sure we plan a safe maneuver around them, no matter their actions. 
We model fully occluded traffic participants (i.e., vehicles, pedestrians,  bikes) through additional occupancy maps, just by adding one more subcategory. This can then be used for motion planning to exert caution. 

More precisely, we represent the occupancy of each class $c \in \mathcal{C}$ as 
a collection of categorical random variables $o^{t,c}_{i,j}$ over space and time. Space is discretized into a BEV spatial grid on the ground plane with a resolution of 0.4 m/pixel, where $i, j$ denotes the spatial location.
Time is discretized into 11 evenly spaced horizons into the future, ranging from 0 to 5 seconds, every 0.5 seconds.

To obtain the output logits $l^{t, c}$ of these spatio-temporal discrete distributions we
employ a multi-scale context fusion by performing two parallel fully convolutional networks with different dilation rates. One stream performs regular 2D
convolutions over $\mathcal{F}_{2x}$, providing very local,
fine-grained features needed to make accurate predictions in the recent
future. The other stream takes the coarser features $\mathcal{F}$ and performs dilated 2D convolutions to obtain a bigger receptive field that is able to place occupancy mass far away from the
initial actor location for those that move fast. We then concatenate the two feature maps into $\mathcal{F}_\text{occ}$. Finally, we design an efficient recurrent occupancy update for each root class to output the logits for all its subclasses 
$$l^{t, c} = l^{t-1, c} + \mathcal{U}^t_\theta(\mathcal{F}_\text{occ} \parallel \mathcal{I}(l^{0:t-1, c}))$$  
where $\mathcal{U}^t_\theta$ is  a neural network that contains a transposed convolution to upsample the resolution by 2, $l^{0:t-1, c}$ are the predicted logits up to timestep $t-1$, $\mathcal{I}$ is a 2x bilinear interpolation, and $\parallel$ represents feature-wise concatenation. 
We perform the recurrence at a lower resolution to reduce the memory impact. We refer the reader to Fig.~\ref{fig:diagram} for a detailed illustration of the recurrency.
Recurrent convolutions provide the right inductive bias to express the intuition that further future horizons need a bigger receptive field, given that actors could have moved away from their starting location.
Finally, to output the categorical distribution $o^{t,c}_{i,j}$ we use a softmax across the mutually-exclusive subclasses of the root class $c$, for each space grid cell ${i, j}$ and time horizon $t$.

\subsection{Motion Planning}
\label{sec:planning}

Given the occupancy predictions $\mathbf{o}$ and the input data $\bx$ in the form of the HD-map, the high level route, traffic-lights states, and the kinematic state of the SDV,  we perform  motion planning by sampling a diverse set of trajectories for the ego-car and pick the one that minimizes a learned cost function as follows: 
\begin{align}
\label{eq:planning}
{\tau}^* = \argmin_{\tau}  f({\tau}, \bx, o; w) 
\end{align}
Here $w$ represents the learnable parameters of the planner. The objective 
function $f$ is composed of subcosts, $f_o$, that make sure the trajectory is safe with
regards to the semantic occupancy forecasts, as well as other subcosts, $f_r$ related to 
comfort, traffic rules and progress in the route (see Fig. \ref{fig:costfunctions}). Thus 
\begin{align}
\label{eq:occ_others}
f({\tau}, \bx, o; w) =   f_o({\tau}, \bx, o; w_o) + f_r({\tau}, \bx, o; w_r)
\end{align}
 with $w=(w_r, w_o)$  the vector of all learnable parameters  for the motion planner. 
We now describe the safety costs in  details, as it is one of our major  contributions. We include a very brief explanation of $f_r$ and refer the reader to the supplementary material for more details.

\medskip\noindent\textbf{Safety Cost:}
The SDV should not collide with other objects on the road
and needs to navigate cautiously when it is uncertain. For this purpose, we
use the predicted semantic-occupancy $o$ to penalize trajectories that
intersect occupied regions. In particular, at each time step $t$ of trajectory $\tau$, we find all the cells in the occupancy layer that have intersection with the SDV polygon (with a safety margin indicated by parameter $\lambda$), and conservatively use the value of the cell with maximum probability as occupancy subcost, denoted by $o_c(\tau, t, \lambda)$. 
Then the safety cost is computed by

\begin{align}
    \label{eq:collision}
    f_o(\tau, o)=\sum_t \sum_c w_c o_c(\tau, t, 0) + w_{cv} o_c(\tau, t, \lambda)v(\tau,t)
\end{align}
with $w_c$ and $w_{cv}$  the weighting parameters.
Note that the first term penalizes trajectories that intersect regions with high
occupancy probability whereas the second term penalizes high-velocity motion in areas with uncertain occupancy.

\begin{figure}[t]
    \centering
    \subfigure[]{\label{fig:collision_cost}\includegraphics[width=14mm]{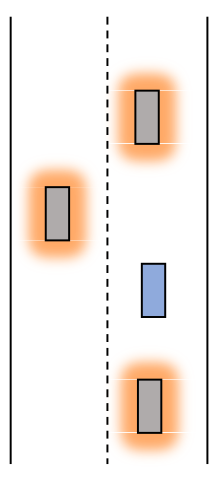}}
    \subfigure[]{\label{fig:drivingpath_cost}\includegraphics[width=14mm]{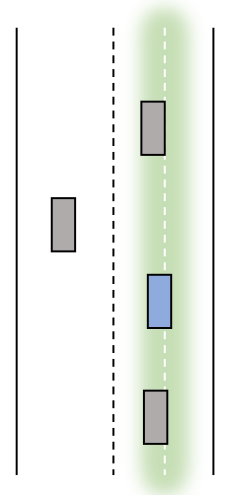}}
    \subfigure[]{\label{fig:boudndary_cost}\includegraphics[width=14mm]{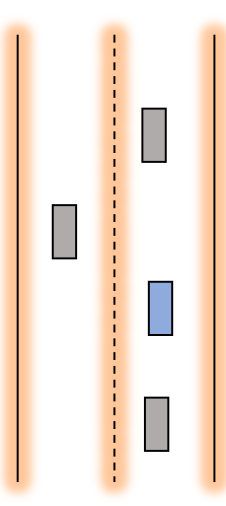}}
    \subfigure[]{\label{fig:trafficlight_cost}\includegraphics[width=17mm]{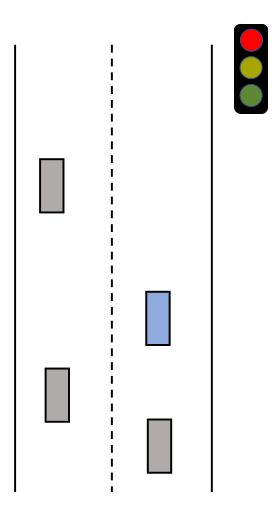}}
    \subfigure[]{\label{fig:dynamics_cost}\includegraphics[width=14mm]{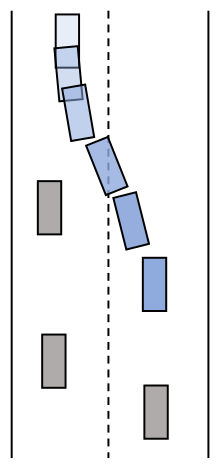}}
    \subfigure[]{\label{fig:route_cost}\includegraphics[width=14mm]{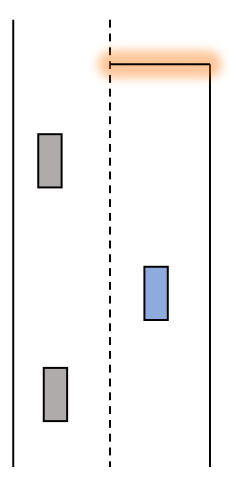}}
    \subfigure[]{\label{fig:progress_cost}\includegraphics[width=13mm]{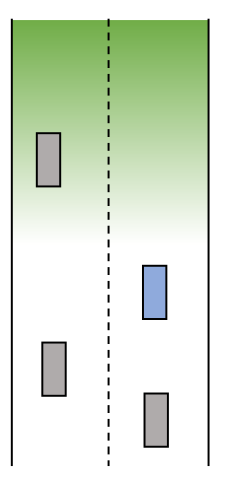}}
    \caption{\textbf{Examples of the motion planner cost functions:} (a) collision, (b) driving-path, (c) lane boundary, (d) traffic-light, (e) comfort, (f) route, (g) progress. }
    \label{fig:costfunctions}
\end{figure}

\medskip\noindent\textbf{Traffic rules, Comfort and Route Progress Costs:}
The trajectory of the SDV must obey  traffic rules. We use the information
available in the map to penalize trajectories that violate the lane
boundaries, road boundaries, stop signs, red traffic-lights, speed-limit, and
do not stay close to the lane center. 
As it is common in self-driving systems, the mission route is given to our planner as a sequence of lanes that the
SDV needs to follow to reach the destination. 
We penalize the number of
lane-changes required to switch to these lanes. This encourages behaviors that
are consistent with the input route. Additionally, in order to promote comfortable driving, we
penalize trajectories for acceleration and violation thereof, lateral
acceleration and violation thereof, jerk and violation thereof, curvature and 
its first and second derivatives. Note that the violations are computed with regards to a predefined threshold
that is considered comfortable. Fig. \ref{fig:costfunctions} shows some of the described cost functions.

\medskip\noindent\textbf{Trajectory Parametrization and Sampling:}
The output of the motion planner is a sequence of vehicle states that describes how the SDV should move within the planning horizon.
At each planning iteration, a set of sampled trajectories are evaluated using the cost function in Eq~\ref{eq:occ_others}, and the one with minimum cost is selected for execution.
It is important that the sampled set, while being small enough to allow real-time computation, cover various maneuvers such as lane-following, lane-changes,
and nudging encroaching objects. Hence, to achieve this efficiently, we choose a sampling approach that is aware of the lane structures. In particular, we follow the trajectory parameterization and sampling procedure proposed in~\cite{sadat2019joint,werling2010optimal}, where trajectories are sampled by combining longitudinal motion and lateral deviations relative to a particular lane (e.g.,~current SDV lane, right lane). Consequently, the sampled trajectories correspond to appropriate lane-based driving with variations in lateral motions which can be applied to many traffic scenarios. The details of the sampling algorithm are presented in the supplementary material.

\subsection{Learning}
\label{sec:learning}
We trained our full model of perception, prediction and planning end-to-end. 
The final goal is to be able to drive safely and comfortably
similar to human demonstrations. Additionally, the model should forecast the
semantic occupancy distributions that are similar to  what happened in the real scene. %
We thus learn the model parameters by exploiting these two loss functions:
\begin{eqnarray}
\label{eq:total_loss}
\mL = \lambdaS\mLS+\lambdaM\mLM
\end{eqnarray}

\medskip\noindent\textbf{Semantic Occupancy Loss:}
This loss is defined as the cross entropy between the ground truth distribution $p$ and the predicted distribution 
$q_\phi$ of the semantic occupancy random variables $o_{i, j}^{t,c}$. 
\begin{eqnarray}
\label{eq:semantic_loss}
\mLS = H(p, q_\theta) = - \sum_t \sum_{c \in \mathcal{C}} \sum_{i, j \in \mathcal{S}^{t,c}} \sum_{o_{i, j}^{t, c} \in \mathcal{O}^c} p(o_{i, j}^{t,c}) \log q_\theta(o_{i, j}^{t,c})
\end{eqnarray}
Due to the highly imbalanced data in terms of spatial occupancy  since the majority of the space is free, 
we obtain the subset of spatial locations $\mathcal{S}^{t,c}$ at time $t$ for class $c$  by performing 
hard negative mining.

\medskip\noindent\textbf{Planning Loss:}
Since selecting the minimum-cost trajectory within a discrete set is not differentiable, we use the max-margin 
loss to penalize trajectories that have small cost and are different from
the human driving trajectory or are unsafe. Let $\bx$ and $\tau_h$ be the input and 
human trajectory respectively for a given example. We utilize the max margin loss to encourage the human 
driving trajectory to have smaller cost $f$ than other trajectories. In particular,
\begin{eqnarray}
\label{eq:maxmargin}
\mLM = \max\limits_{\tau} \bigg[~ f_{r}(\bx, \tau_{h})-f_{r}(\bx, \tau) +l_\text{im} + \sum_{t}\big[f^{t}_{o}(\bx, \tau_{h})-f^{t}_{o}(\bx, \tau) +l^{t}_\text{o} \big]_+ \bigg]_+
\end{eqnarray}
where $f^t_o$ is the occupancy cost function at time step $t$, $f_r$ is the rest of 
the planning subcosts as defined in Section \ref{sec:planning} (note that we omitted $o$ and $w$ from $f$ for brevity), and $[]_+$ represents the ReLU function.
The imitation task-loss $l_\text{im}$ 
measures  the $\ell_1$ distance  between trajectory $\tau$ and the ground-truth for the entire horizon,
and the safety task-loss $l^t_\text{o}$ accounts for collisions and their severity at each trajectory step.

\section{Experimental Evaluation} 

\medskip\noindent\textbf{Dataset and Training:}
We train our models using our large-scale dataset that includes challenging scenarios where the operators are instructed to drive smoothly and in a safe manner. It contains 6100 scenarios for the  training set, while validation and test sets contain 500 and 1500 scenarios. Each scenario is 25 seconds. Compared to KITTI\cite{geiger2012we}, our dataset has 33x more hours of driving and 42x more objects. 
We use exponentiated gradient decent to update the planner parameters and Adam optimizer for the occupancy forecasting.
We scale the gradient that is passed to perception and prediction from the planner to avoid instability in P\&P training.

\medskip\noindent\textbf{Baselines:}
We compare against  the following baselines: \textbf{ACC} which  performs a simple car-following behavior using the measured state of the lead vehicle. \textbf{Imitation Learning (IL)} where the future positions of the SDV are predicted directly from the fused LiDAR and map features (Fig \ref{fig:diagram}), and is trained using L2 loss. \textbf{NMP} \cite{zeng2019neural} where a planning cost-map is predicted from the fused features directly and detection and predictions are treated only as an axillary task. \textbf{PLT}: which is the joint behavior-trajectory planning method of \cite{sadat2019joint}, where planning is accomplished using a combination of interpretable subcosts, including collision costs with regards to predicted trajectories of actors. However, the detection and prediction modules are trained separately from the planner.

\medskip\noindent\textbf{Metrics:}
Planning metrics include \textbf{cumulative collision rate} indicating the percentage of collisions with ground-truth bounding-boxes of the actors at each trajectory time step, \textbf{L2 distance} to human trajectory which indicates how well the model imitated the human driving, \textbf{jerk} and \textbf{lateral acceleration} which show how comfortable the produced trajectories are. We also measure the \textbf{progress} of the SDV along the route.

\medskip\noindent\textbf{Results:}
The first set of experiments are performed in an open-loop setting in which the LiDAR data up to the current timestamp is passed to the model and the generated trajectory is assumed to be executed by the ego vehicle for the 5sec planning horizon (as opposed to closed-loop execution where the trajectory is constantly replanned as new sensor data becomes available). Table \ref{table:main} shows the planing metrics for our proposed method and the baselines. It shows that our proposed model (P3) outperforms all the baselines in (almost) all planning metrics. In particular, our motion planner generates much safer trajectories, with 40\% less collisions at 5s compared to PLT. It also outperforms NMP by a very significant margin, which could be due to our consistent use of perception and prediction outputs in motion planning, as opposed to the free-form cost volume from sensor data in  \cite{zeng2019neural}. Another aspect that we observed to improve safety was the temporally smoother occupancies output by our recurrent occupancy update as opposed to a convolutional one. A more nuanced detail that could also contribute to our increased safety is the pooling of the cost on the space occupied by the SDV, as opposed to the simple indexing on its centroid previously proposed by \cite{zeng2019neural}. Our model also produces less jerk which indicates the effectiveness of including multiple interpretable subcosts in the planning objective. Besides, our model exhibits much closer behavior to human compared to IL that has been optimized to match human trajectories. The progress metric also shows that our model is less agggressive compared to the other baselines and similar to IL.
\begin{table}[t]
	\centering
	\resizebox{\columnwidth}{!}{%
		\scriptsize
		\begin{threeparttable}
			\begin{tabularx}{\textwidth}{
				>{\centering\arraybackslash}c |  %
				>{\centering\arraybackslash}X >{\centering\arraybackslash}X >{\centering\arraybackslash}X |
				>{\centering\arraybackslash}X >{\centering\arraybackslash}X >{\centering\arraybackslash}X |
				>{\centering\arraybackslash}X | >{\centering\arraybackslash}X | >{\centering\arraybackslash}X }
			\toprule
			Model  	                            &  \multicolumn{3}{c|}{Collisions rate (\%)} &\multicolumn{3}{c|}{L2 human}      & Jerk  & Lat. acc. & Progress  \\
			{}  	                            & 1s  	& 3s &  5s &@1s &  @3s	 &  @5s	 &       &   &	    \\
			\midrule
			ACC 	                            &   0.31    	&   2.00    &  8.73   &	0.20 &  1.75  	 &	 5.16	 &   1.74    &  2.87  & 29.3			  \\
			IL	 	 			                &   1.47 &   4.33 &  12.29& 0.33&  1.46	 &  3.37 &   -    &  - & 27.5  			  \\
			NMP\cite{zeng2019neural} 					            &   0.17 &  0.72 &  5.22& 0.23&  2.19 	 &	5.61 &   4.36 &  \textbf{2.86} & 31.7    \\
			PLT\cite{sadat2019joint}            &   0.07 &   0.40 &   2.94& \textbf{0.18}&  1.35	 &	3.80 &   1.52 &  3.03 & 28.0   		  \\
			\midrule
			P3 (Ours) 					            &   \textbf{0.05} &  \textbf{0.17}&   \textbf{1.78}& \textbf{0.18}&  \textbf{1.18}   &	\textbf{3.34}	 &   \textbf{1.27}    &  2.89 & 27.6   \\
			\bottomrule
			\end{tabularx}
		\end{threeparttable}
	}
	\caption{\textbf{Comparison against other methods}}
	\label{table:main}
\end{table}

\begin{table}[t]
	\centering
	\resizebox{\columnwidth}{!}{%
		\scriptsize

		\begin{threeparttable}
			\begin{tabularx}{\textwidth}{
				>{\centering\arraybackslash}c |  %
				>{\centering\arraybackslash}X >{\centering\arraybackslash}X >{\centering\arraybackslash}X |
				>{\centering\arraybackslash}X >{\centering\arraybackslash}X >{\centering\arraybackslash}X |
				>{\centering\arraybackslash}X >{\centering\arraybackslash}X >{\centering\arraybackslash}X |
				>{\centering\arraybackslash}X | >{\centering\arraybackslash}X | >{\centering\arraybackslash}X}
			\toprule
			ID  	              & \multicolumn{1}{c}{e2e}& \multicolumn{2}{c}{Prediction}             &  \multicolumn{3}{|c|}{Collision rate (\%)} &\multicolumn{3}{c|}{L2 human}      & Jerk  & Lat. acc. & Progress \\
			{}  	              & & Traj.& Occup.          & 1s  & 3s & 5s &@1s &  @3s	 &  @5s	 &       &  	&    \\
			\midrule
			$\mathcal{M}_1$ 					  &  &\checkmark &   		 &0.07& 0.40& 2.94& \textbf{0.18}& 1.35&	3.80& 1.52 & 3.03 & 28.0 \\
			$\mathcal{M}_2$ 					  &\checkmark  & \checkmark& &\textbf{0.05}& 0.32& 2.21& \textbf{0.18}& 1.35& 3.65& 1.50 & 2.85 & 27.8\\
			$\mathcal{M}_3$ 					  & &\checkmark  & \checkmark &\textbf{0.05}& 0.22& 2.36& \textbf{0.18}& 1.27& 3.64& 1.38 & 2.93 & 28.0 \\
			$\mathcal{M}_4$ 					  &  & & \checkmark  		 &\textbf{0.05}& 0.20& 1.96& \textbf{0.18}& 1.21& 3.49& \textbf{1.23} & \textbf{2.78} & 27.3  \\
			$\mathcal{M}_5$ 					  & \checkmark & & \checkmark  &   \textbf{0.05} &  \textbf{0.17} &  \textbf{1.78}& \textbf{0.18}&  \textbf{1.18}   &	\textbf{3.34}	 &   1.27    &  2.89  & 27.6 \\

			\bottomrule
			\end{tabularx}

		\end{threeparttable}
	}
	\caption{\textbf{Ablation study}}
	\label{table:ablation}
\end{table}

\medskip\noindent\textbf{Ablation Study:}
We report the result of the ablation study in Table \ref{table:ablation}. Our best model is $\mathcal{M}_5$ (P3 in Table~\ref{table:main}) corresponds to the semantic occupancy and the motion planner being jointly trained. $\mathcal{M}_1$ and $\mathcal{M}_2$ perform detection and multi-modal trajectory prediction which is used in motion planner to form collision costs. Overall, end-to-end training of perception and planning modules improve safety as indicated by the collision metrics. Furthermore, using occupancy representation yields much better performance in driving metrics. The progress metric also indicates that the occupancy model is not overly cautious and the advancement in the route is similar to other models.
Note that we also include $\mathcal{M}_3$ which is similar to $\mathcal{M}_1$, but the predicted trajectories are rasterized to form an occupancy representation for motion planning.

\begin{figure}[t]
    \centering     %
    \subfigure[t=0s]{\label{fig:a}\includegraphics[width=40mm]{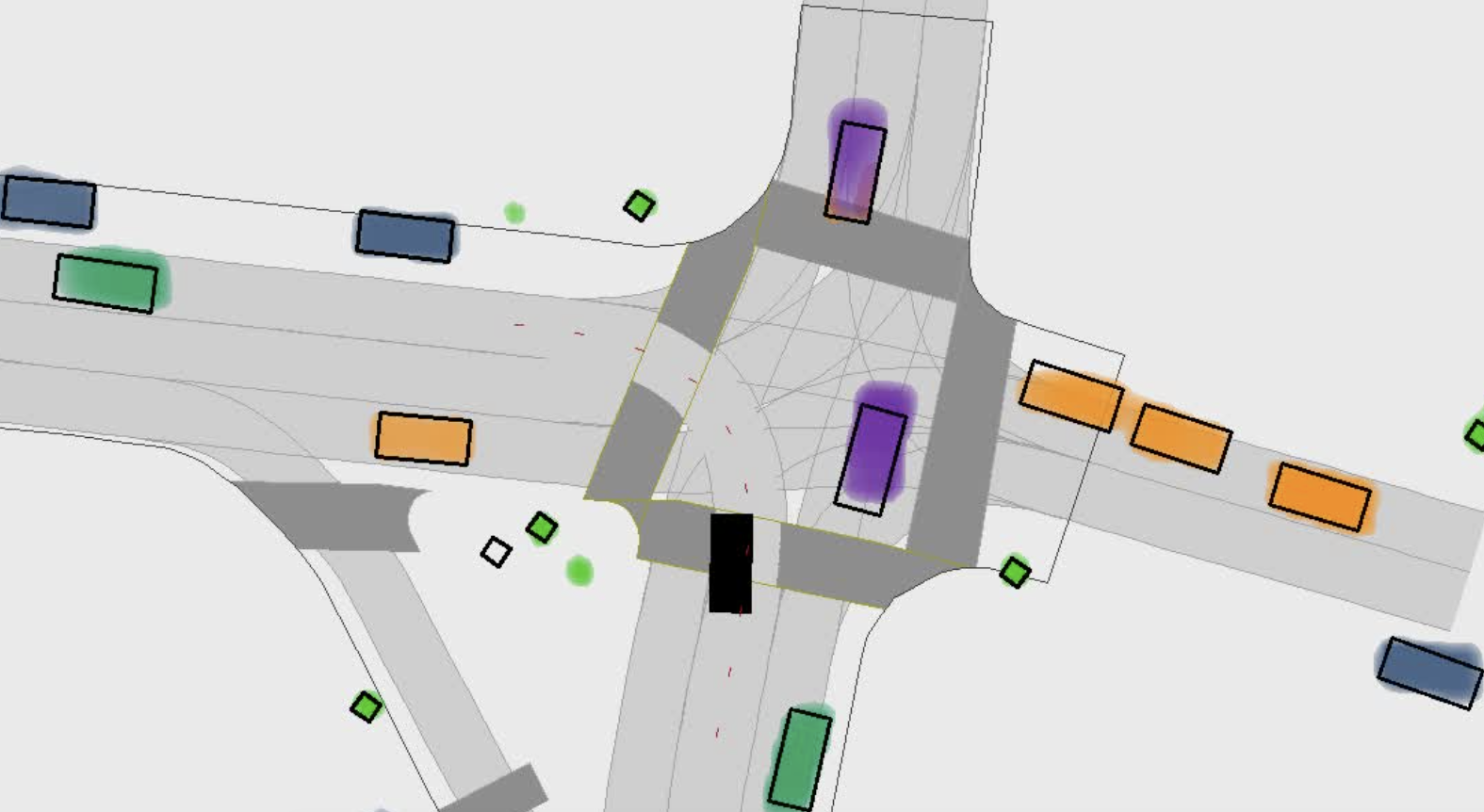}}
    \subfigure[t=2.5s]{\label{fig:b}\includegraphics[width=37mm]{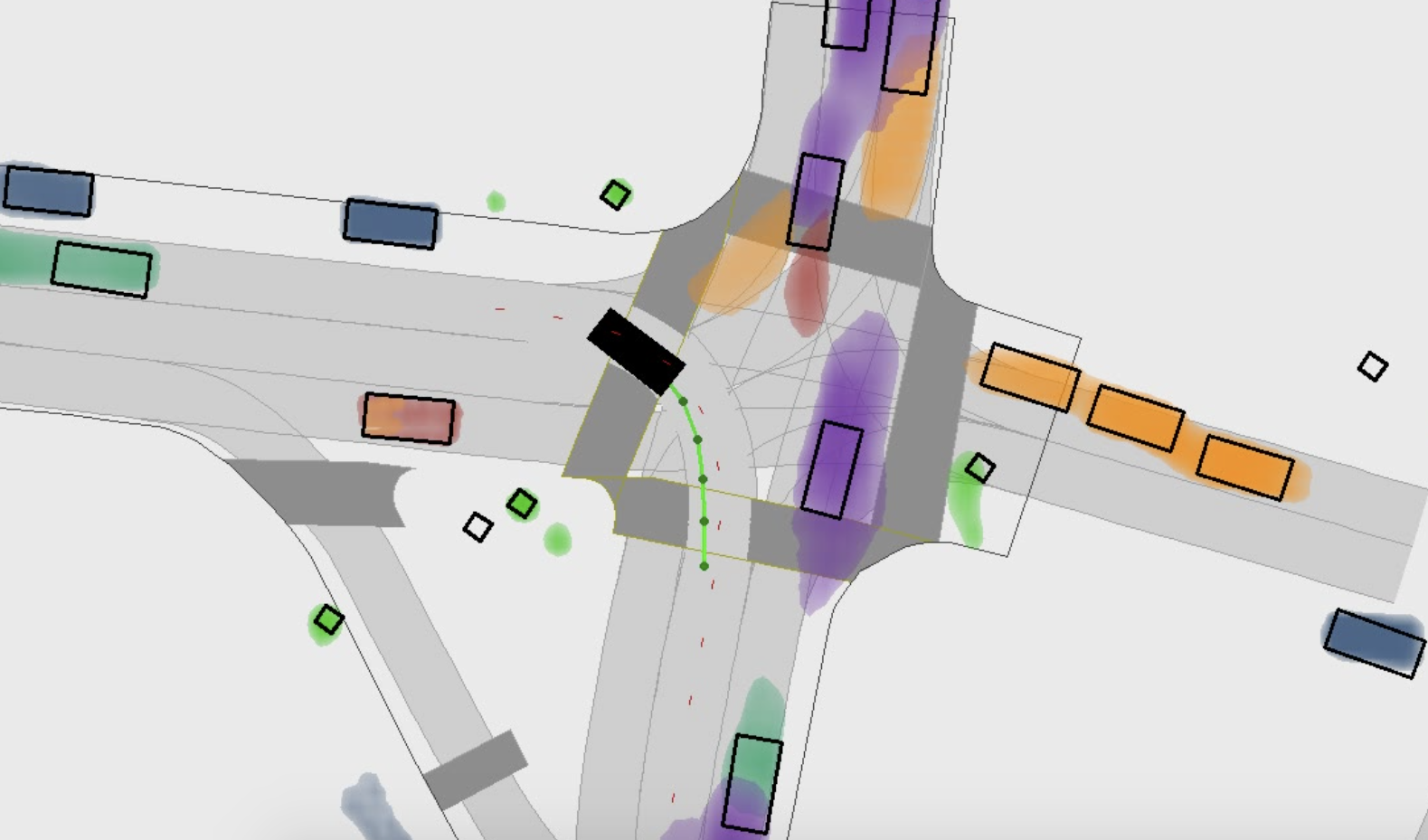}}
    \subfigure[t=5s]{\label{fig:c}\includegraphics[width=40mm]{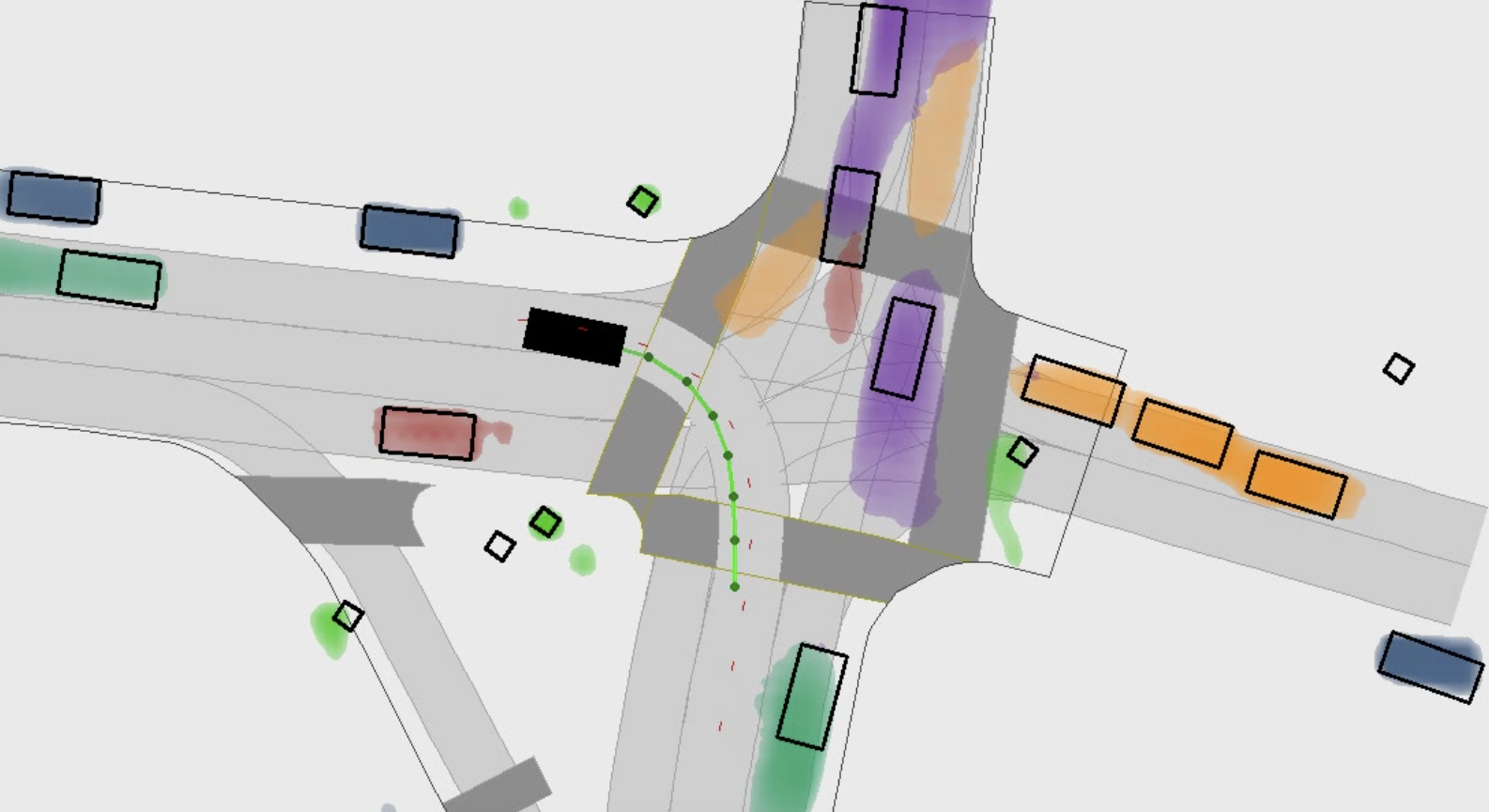}}
    \subfigure[t=0s]{\label{fig:d}\includegraphics[width=39mm]{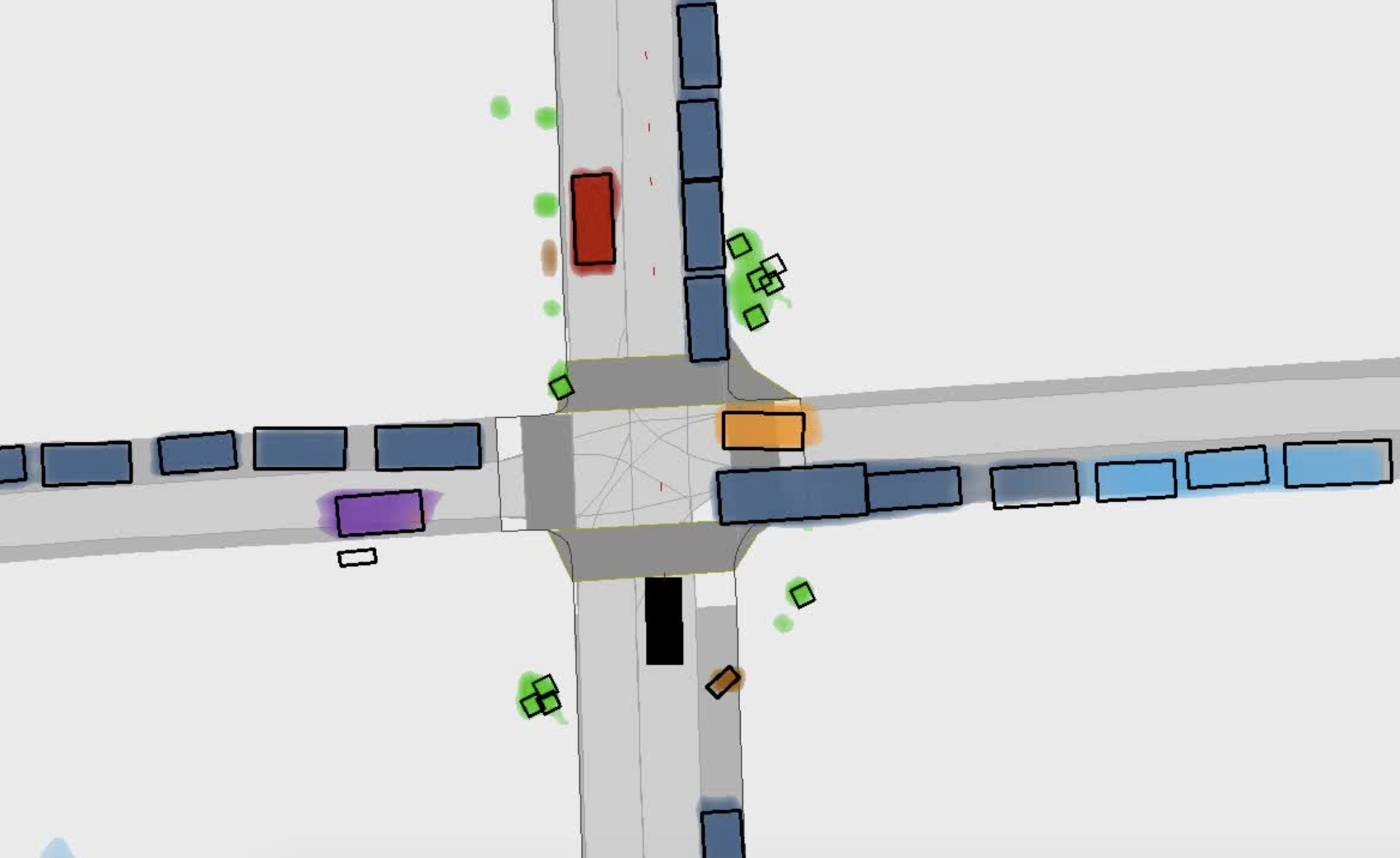}}
    \subfigure[t=2.5s]{\label{fig:e}\includegraphics[width=40mm]{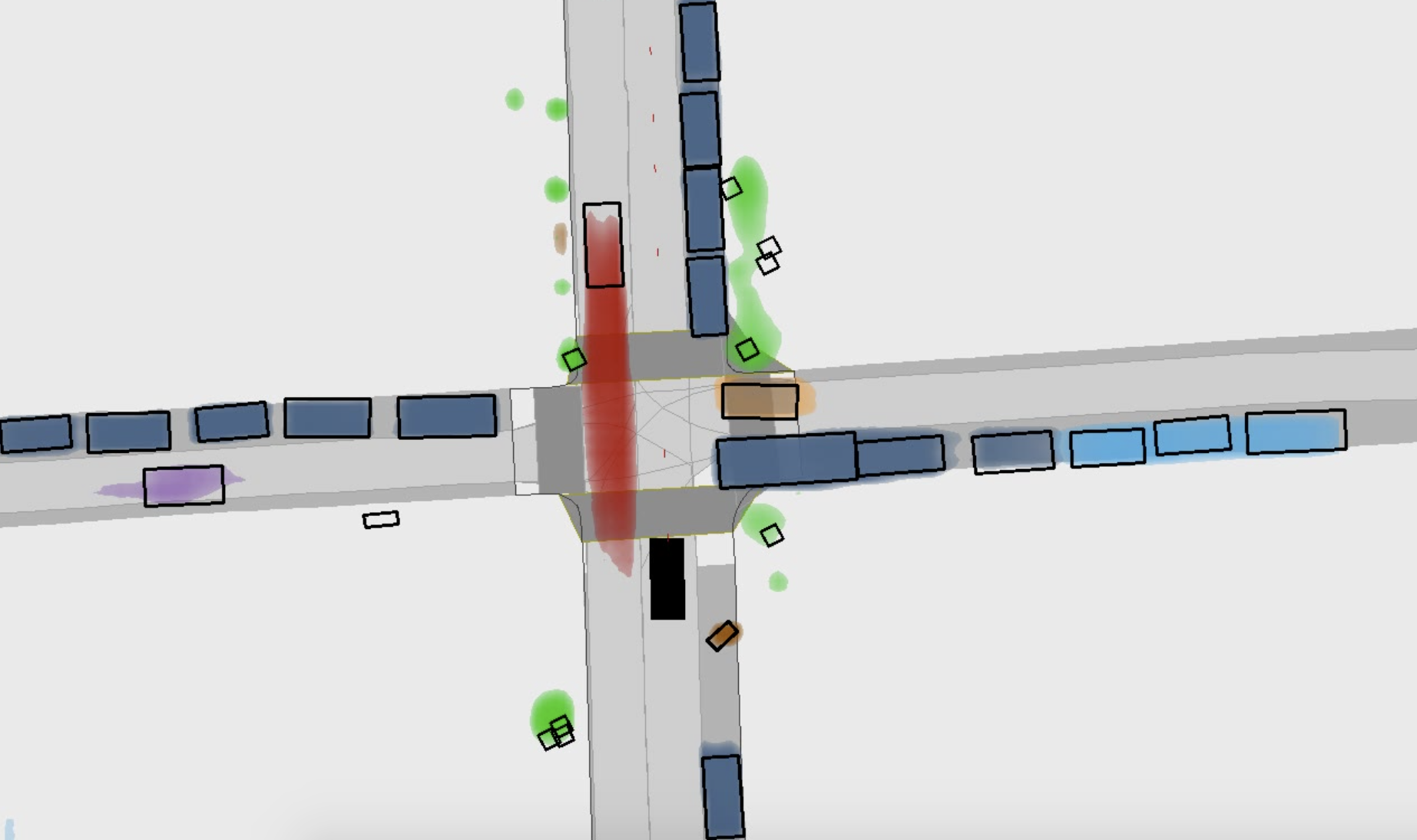}}
    \subfigure[t=5s]{\label{fig:f}\includegraphics[width=39mm]{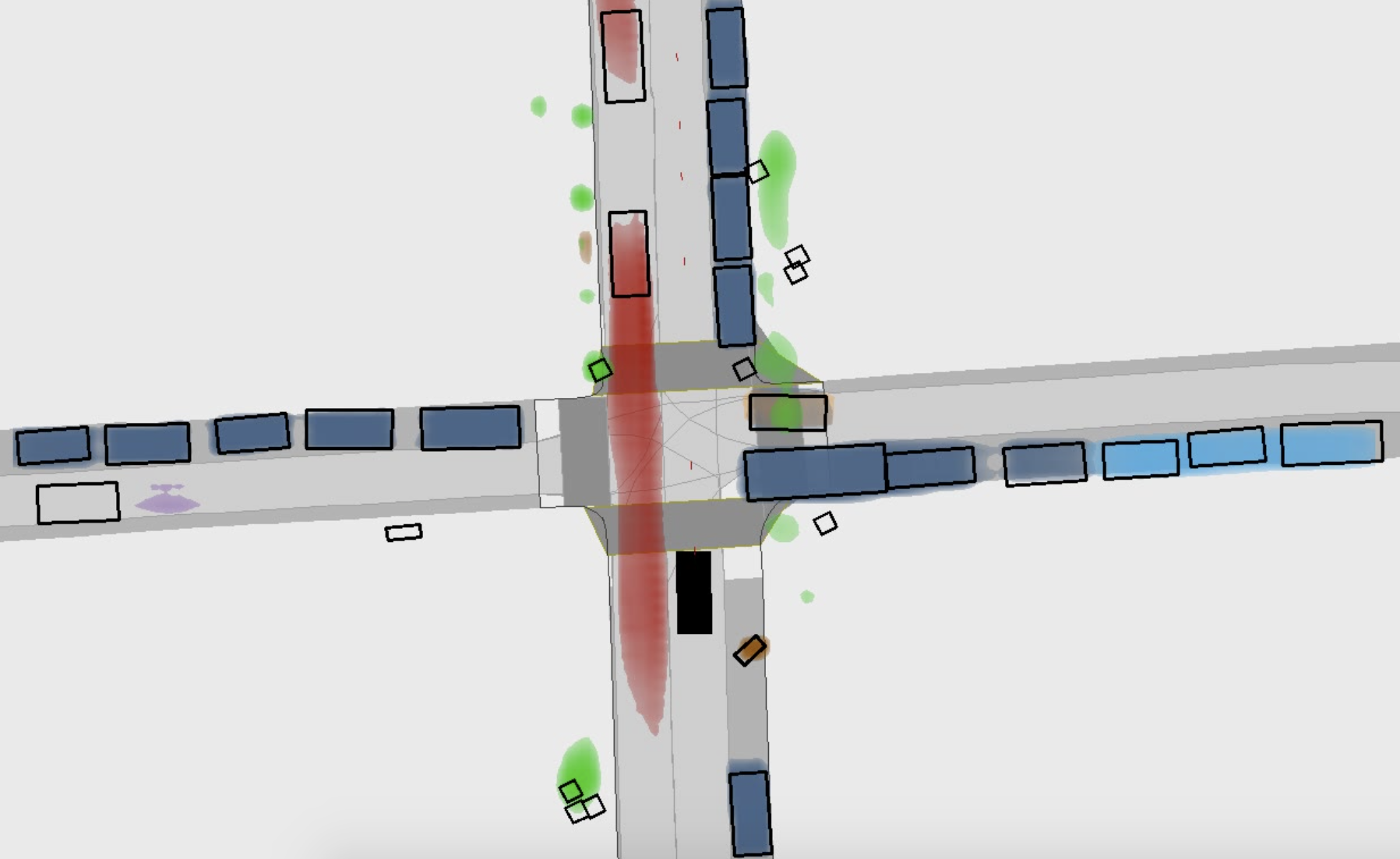}}
    \caption{\textbf{Qualitative results}: The colors red, orange, dark-green, dark-blue, and purple respectively shows vehicle with sub-categories of oncoming, conflicting, on-route, stationary and others. Pedestrians and bicyclists are shown with light green and brown colors. Cyan color is used to show occlusion for all classes. We also show the ground-truth bounding boxes of all actors, and planned trajectory of the ego vehicle (solid black rectangle)}
    \label{fig:qualitative}
\end{figure}

\medskip\noindent\textbf{Qualitative results:}
Fig.  \ref{fig:qualitative} shows examples of generated semantic occupancy layers at different time horizons for two traffic scenarios (refer to the caption for corresponding color of each semantic class). In Fig. \ref{fig:c}, for example, we can see  multiple modes in the prediction of a vehicle with corresponding semantics of conflicting and oncoming. In the bottom scene, our model is able to recognize the occluded region on the right end of the intersection. Furthermore, the oncoming vehicle (red color) which has a low initial velocity is predicted with large uncertainty which is visible in Fig. \ref{fig:f}.

\begin{figure}[t]
    \centering     %
    \subfigure[Scenarios]{\label{fig:closedloopsetup}\includegraphics[width=28mm]{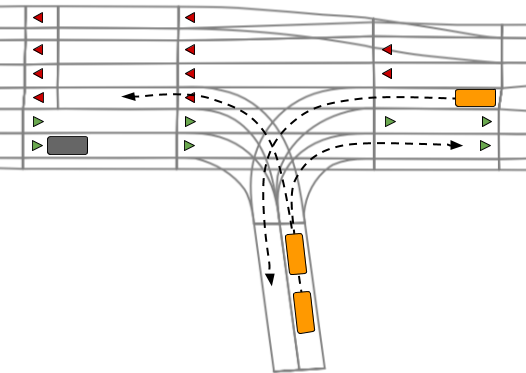}}
    \subfigure[t=0s]{\label{fig:closedloop1}\includegraphics[width=28mm]{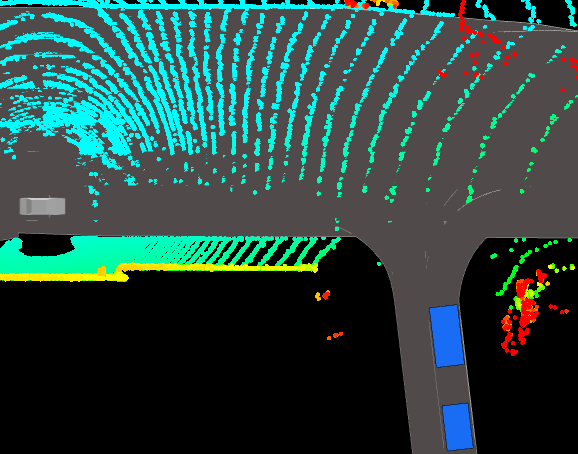}}
    \subfigure[t=5s]{\label{fig:closedloop2}\includegraphics[width=31mm]{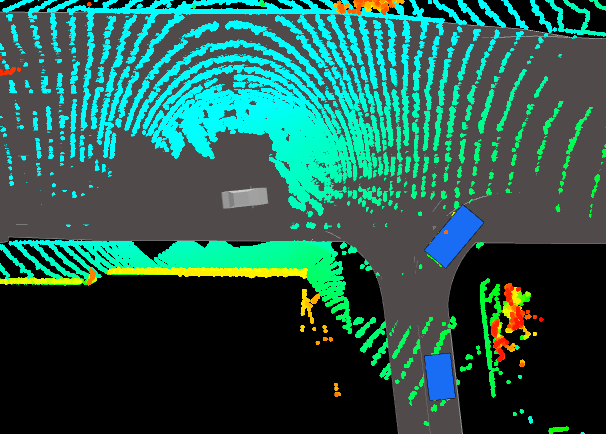}}
    \subfigure[t=10s]{\label{fig:closedloop3}\includegraphics[width=30mm]{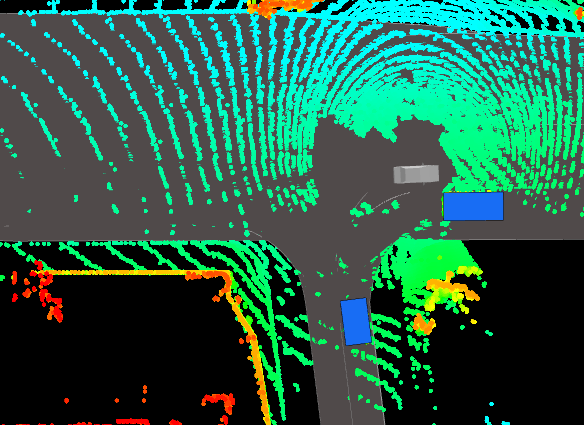}}
    \caption{\textbf{Closed-loop scenario:} The general setup of the closed-loop evaluation is shown in (a) where the SDV (gray vehicle) is approaching an intersection with 3 potential non-reactive vehicles (orange colored) with conflicting routes. The variations of the scenario are generated by including 1 or 2 of the indicated vehicles with various initial speed and location. (b,c,d) show and example of the simulation run at different time horizons.}
    \label{fig:closedloop}
\end{figure}

\begin{table}[t]
	\centering
	\resizebox{\columnwidth}{!}{%
		\scriptsize
		\begin{threeparttable}
			\begin{tabularx}{\textwidth}{
				>{\centering\arraybackslash}c |  %
				>{\centering\arraybackslash}X |
				>{\centering\arraybackslash}X >{\centering\arraybackslash}X >{\centering\arraybackslash}X >{\centering\arraybackslash}X >{\centering\arraybackslash}X}
			\toprule
			ID  	              &  Collision rate (\%) & Jerk  & Lateral acceleration & Acceleration & Deceleration & Speed \\
			\midrule
			$\mathcal{M}_2$ 					  &18.5&4.08&0.24&0.94&-0.79&9.1\\
			$\mathcal{M}_5$ 					  &\textbf{9.8}&\textbf{1.85}&0.17&0.50&-0.50&8.6\\
			\bottomrule
			\end{tabularx}
		\end{threeparttable}
	}
	\caption{\textbf{Closed-loop Evaluation Results} See Table \ref{table:ablation} for definition of $\mathcal{M}_2$ and $\mathcal{M}_5$. The collision rate shows the percentage of the simulation runs where the SDV had collision with other actors. The rest of the metrics show the mean value over all the simulation steps.}
	\label{table:closedloop}
\end{table}
\medskip\noindent\textbf{Closed-loop Evaluation:}
We also perform experiments in a closed-loop simulated environment leveraging realistic LiDAR simulation \cite{siva2020lidarsim}. At each simulation time-step, the simulated LiDAR point-cloud is passed to our model and a trajectory is planned for the ego vehicle and is executed by the simulation for 100ms. This process continues iteratively for 15s of simulation. We tested our models in a scenario with one or two initially-occluded non-compliant actors with trajectories that are in conflict with the route of the ego-vehicle (see Fig \ref{fig:closedloopsetup}). By varying the initial velocity and along-the-lane location of each actor, we created 80 highly challenging traffic scenes (12k frames) for our tests. We compared the performance of our proposed end-to-end autonomy system that uses semantic occupancy with the alternative trajectory-based method ($\mathcal{M}_5$ and $\mathcal{M}_2$ respectively). As   shown in Table \ref{table:closedloop}, our full approach can react safely to the non-compliant vehicles,  resulting in less collisions than $\mathcal{M}_2$. This cautious behavior is also reflected in the rest of the metrics such as jerk, acceleration, and velocity where $\mathcal{M}_2$ exhibits more aggressive behavior.
Fig. \ref{fig:closedloop} demonstrates an example run of the simulation. As the SDV approaches the intersection, the non-reactive vehicle, which is turning right, becomes visible (Fig. \ref{fig:closedloop2}). The planner generates a lane-change trajectory to avoid the slow-moving vehicle (Fig. \ref{fig:closedloop3}).

\section{Conclusion}
In this paper, we have proposed an end-to-end perception, prediction and motion planning model that generates safe trajectories for the SDV from raw sensor data. Importantly, our model not only produces interpretable intermediate representations, but also the generated ego-vehicle trajectories are consistent with the perception and prediction outputs. Furthermore, unlike most previous approaches that employ thresholded activations in detection and trajectory prediction of objects, we use semantic occupancy layers that are able to carry information about low probability objects to the motion planning module. Our experiments on a large dataset of challenging scenarios and closed-loop simulations showed that the proposed method, while exhibiting human-like driving behavior, is significantly safer than the state-of-the-art learnable planners.

\clearpage

\bibliographystyle{splncs}
\bibliography{ref}

\clearpage

\appendix

\vspace*{0.2cm}
{ \noindent \Large \textbf{Supplementary Material}} \\

In this supplementary material, we present the trajectory parameterization and sampling procedure in more details. Additionally, we give an overview of all the planning cost-functions. Further details about training our models are included as well as more qualitative results.

\section{Trajectory Parametrization and Sampling}
\begin{figure*}
    \centering
    \includegraphics[width=0.8\textwidth]{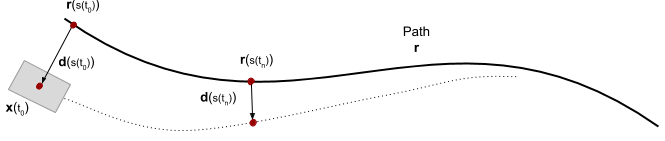}
    \caption{\textbf{Frenet Frame}}
    \label{fig:frenet}
\end{figure*}

The output of the planner is a trajectory that consists of a sequence
of bicycle model states $\tau={p_t}$, $p_t=(x,y,\theta,v,\kappa,a)$, where $(x,y)$ is the position of the center of the rear axle of the vehicle, $\theta$ is the heading, $v$ and $a$ are the forward velocity and acceleration, and $\kappa$ is the curvature of the vehicle path which can be converted to steering angle. Candidate trajectories can be generated by sampling various curvature and acceleration values and using the kinematic bicycle model to obtain the other states (position, heading, velocity) \cite{kong2015kinematic}. However, this approach will be very inefficient as most of the sampled trajectories will not exhibit proper lane-based driving. Therefore we adopt an alternative approach which uses the lanes structures to generate higher quality trajectories. Specifically, we sample trajectories in \textit{Frenet Frame} of the driving-path of the desired lane \cite{werling2010optimal}, i.e.~instead of kinematic bicycle-model state, we use longitudinal position and lateral offset (and their higher order derivatives) relative to a driving-path to represent a trajectory. Figure \ref{fig:frenet} demonstrates such parametrization, where $r$ is the driving path parametrized by arc-length $s$, $s(t)$ is the longitudinal position of the vehicle parametrized by time $t$, and $d(s)$ is the lateral offset from the driving-path parametrized by arc-length $s$. Each pair of $s(t)$ and $d(s)$ can describe a bicycle model trajectory. 
Specifically, the Frenet state defined as $[s,\dot{s},\ddot{s}, d, d^{\prime}, d^{\prime\prime}]$ can  be transformed to bicycle model state $(x,y,\theta,v,\kappa,a)$ \cite{werling2010optimal}. Note that $\dot{(.)}:=\frac{\partial}{\partial t}$, and $(.)^\prime:=\frac{\partial}{\partial s}$ denote the derivatives with respect to time and arc-length.

Our trajectory sampling procedure is as follows: we first sample a set of longitudinal trajectories that convert various velocity profiles such as stopping, accelerating to a specific-velocity, maintaining the current velocity. Then, for each longitudinal trajectory, we sample lateral trajectories that include maneuvers such as nudging, changing-lane, and following the driving-path. Combining the two sets results in bicycle model trajectories that are proper lane-based trajectories including lateral maneuver variations to handle challenging scenarios.

We use quintic and quartic polynomials to represent longitudinal and lateral trajectories. Specifically, the set of longitudinal trajectories $\mathcal{S}=\{s(t)\}$ are sampled by using a large set of mid-conditions $[\dot{s}(t_1),t_1]$ and end-conditions $[\dot{s}(T),T]$ and solving for two quartic polynomials that are stitched together. The acceleration ($\ddot{s}$) at $t_1$ and $T$ are fixed at 0. We parameterize lateral trajectories $[d(s), d^\prime(s), d^{\prime\prime}(s)]$ in terms of the longitudinal distance $s$.
We generate a set of mid-conditions $[d(s_1),s_1]$ and  fix $d^\prime(s_1)$ and $d^{\prime\prime}(s_1)$ to be 0. We require the lateral trajectories to approach the driving path and hence the end-conditions $[0, 0, 0]$. The initial, mid- and end-conditions are used to obtain two quintic polynomials that specify the lateral offsets for each longitudinal trajectory. 
Each pair of sampled longitudinal and lateral trajectories $[s(t),d(s)]$ are transformed back to a bicycle model trajectory.

\section{Motion Planner Cost Functions}
In this section we present the details of the cost functions that are used to evaluate trajectories in the motion planner. Figure \ref{fig:costfunctions} shows a subset of the subcosts.
\begin{figure}[t]
    \centering
    \subfigure[Collision]{\label{fig:collision_cost}\includegraphics[width=21mm]{images/c_obstacle}}
    \subfigure[Headway]{\label{fig:headway_cost}\includegraphics[width=23mm]{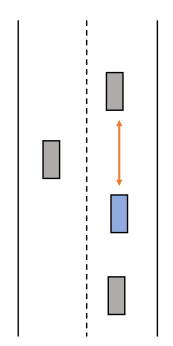}}
    \subfigure[Path]{\label{fig:drivingpath_cost}\includegraphics[width=21mm]{images/c_drivingpath}}
    \subfigure[Lane]{\label{fig:boudndary_cost}\includegraphics[width=21mm]{images/c_boundary}}
    
    \subfigure[Traffic lights]{\label{fig:trafficlight_cost}\includegraphics[width=25mm]{images/c_trafficlights}}
    \subfigure[Comfort]{\label{fig:dynamics_cost}\includegraphics[width=20mm]{images/c_dynamics}}
    \subfigure[Route]{\label{fig:route_cost}\includegraphics[width=20mm]{images/c_route}}
    \subfigure[Progress]{\label{fig:progress_cost}\includegraphics[width=19mm]{images/c_progress}}
    \caption{\textbf{Examples of the motion planner cost functions} }
    \label{fig:costfunctions}
\end{figure}

\subsubsection{Collision, Safety-margin, and Headway:}
The trajectory of the SDV should be collision-free and at a safe distance from surrounding objects. We use collision and safety-distance costs (Fig. \ref{fig:collision_cost}) to penalize trajectories that have spatio-temporal overlap with the predicted trajectories of other actors or violate a safety margin. This is achieved by computing the distance between SDV polygon and the predicted polygon of all the other actors at each timestep. Furthermore, the SDV should maintain a safe headway to the leading vehicle, such that if the lead vehicle applies a hard break, the SDV can slow-down smoothly to avoid collision and uncomfortable breaking (Fig. \ref{fig:headway_cost}). This cost is computed using the relative longitudinal distance of the SDV and the lead vehicle as well as their velocities. Note that the above costs are defined when the prediction module generates bounding-boxes and trajectories for the actors. The collision and safety subcosts using occupancy representations are described in the paper. Also, if the prediction module forecasts multiple modes of trajectories for each actor, the above subcosts are computed for each predicted trajectory and are weighted by the probability of that trajectory mode. 

\subsubsection{Driving-path, Lane and Road Boundaries:}
The SDV should adhere to the structure of the lanes and roads. For example, it is expected that vehicles stay close to the center of the lane and not move over the boundaries of the lane and roads. For this purpose, we introduce subcosts that penalize the violation of lane and road boundaries as well as the distance to the driving-path of the lane. These subcosts are demonstrated in Fig. \ref{fig:drivingpath_cost} and \ref{fig:boudndary_cost}. 

\subsubsection{Speed-limit, Traffic Lights, and Stop Signs:}
We penalize the violation of speed-limit at each trajectory step to promote driving at the regulated speed-limit. Furthermore, for each red traffic-light or stop sign, the SDV needs to come to a stop at the intersection stop-line, represented by a longitudinal position along the lane. Therefore, we introduce cost functions where the violation of each stop-line by a trajectory is penalized. 

\subsubsection{Route, Progress, and Cost-to-go:}
The SDV is given a high-level route represented as a sequence of lanes. Although the SDV can use other lanes that are adjacent to the route lanes, we penalize the number of lane-changes that is required to return back to the route lanes. Furthermore, if the SDV is using a dead-end lane (i.e., lanes that diverge from the route), we penalize violation of a distance-threshold to the end of that lane such that the SDV is forced to change the lane (Fig. \ref{fig:route_cost}). Trajectories are also rewarded (negative cost) by the distance they move along the lane to promote progress in the route (Fig. \ref{fig:progress_cost}).

We also introduce a cost that captures what comes beyond the planning horizon. Specifically, for each trajectory we penalize the deceleration that is needed to reduce the SDV speed, from the value specified by the last trajectory point to an acceptable lower value, due to upcoming speed-limits, stop-signs, or red traffic light.

\subsubsection{Dynamics and Comfort:}
We prune trajectories that violate vehicle constraints such as maximum acceleration or curvature to only allow executable trajectories for the SDV.
Furthermore, since rapid changes in acceleration or steering lead to uncomfortable rides, we penalize such aggressive motions. Specifically, we penalize jerk and violation thereof, acceleration and violation thereof, lateral acceleration and violation thereof, curvature and its first and second derivatives. All the violations above are computed based on a predefined threshold.

\section{Training details}
\subsubsection{Optimizer:}
We use Adam optimizer to update the weights in the perception backbone and occupancy forecasting networks, with a base learning rate of $1e-5$. We use exponentiated gradient descent to optimize the planning parameters such that the subcosts' weights remain greater than zero after each iteration $i$, with a planning base learning rate $\alpha=1e-3$:
$$w^{(i+1)}=w^{(i)} \exp (-\alpha \mathbf{g})$$
Here $\alpha$ is the learning rate and $\mathbf{g}$ denotes the gradient of $w$.
We employ linear scaling to both learning rates with respect to the batch size. 

\subsubsection{Hyperparameters:}
In our multi-task learning setting, we use a weight of $\lambdaS=1$ for the semantic occupancy cross entropy loss and $\lambdaM=1e-3$ for the motion planning max-margin loss. In the semantic occupancy cross entropy we employ hard negative mining with a ratio of 10 negative examples for each positive one. Note that originally the classification problem is much more imbalanced, with the vast majority of the grid cells being not occupied. More precisely, we first define a subset of negative pixels $\text{Neg}^{t,c}$ over time $t$ and classes $c$, which include a random 10\% of the non-occupied grid cells. Then we pick all the positive examples $\text{Pos}^{t,c}$ and the hardest $10 \cdot |\text{Pos}^{t,c}|$ from $\text{Neg}^{t,c}$ (the ones with the highest loss). Finally we combine the positives and the subset of negatives to from the final subset of pixels $\mathcal{S}^{t,c}$.

\section{Architecture details}
In this section we give more details about the architecture of the recurrent occupancy updates. The backbone network was already explained in details in the paper.
\subsubsection{Recurrent Occupancy Update:} We employ a multi-scale context fusion by performing two parallel fully convolutional networks with different dilation rates. One stream performs regular 2d convolutions over $\mathcal{F}_{2x}$ with a 2-layer CNN with dilation of 1, using 128 feature channels. The other stream takes the coarser features $\mathcal{F}$ and processes it with another 2-layer CNN with dilation of 2 at both layers, using 128 feature channels. We then apply bilinear interpolation to the feature tensor at 2x downsampling to bring it to the lower resolution (4x downsampling), and concatenate these two tensors along the channel dimension to obtain the context $\mathcal{F}_\text{occ}$. Our approach then predicts the occupancy over time in a recurrent fashion, from the context. Note that the context $\mathcal{F}_\text{occ}$ is downsampled 4 times from the input (0.8 m/pixel), but this is too coarse for motion planning (e.g., when trying to turn into tight spaces, it could look like the space is occupied by a parked car when it is not, just due to discretization). Thus, we seek to produce the occupancies at 0.4 m/pixel. Our proposed recurrence looks as follows:
$$l^{t, c} = l^{t-1, c} + \mathcal{U}_\theta(\mathcal{F}_\text{occ} \otimes \mathcal{I}(l^{0:t-1, c}))$$ 
$l^{t, c}$ are the logits for root class $c$ at timestep $t$ into the future. $\mathcal{I}$ is a 2x bilinear interpolation to bring the previous occupancy logits into a resolution of 0.8 m/pixel. $\otimes$ represents feature-wise concatenation. $\mathcal{U}_\theta$ is 2-layer CNN with a hidden dimension of 256, where the first convolution is transposed to upsample the resolution by 2, and the second layer is a regular convolution. The initial occupancy at t=0 $l^{0, c}$ is predicted by a small 2-layer CNN from $\mathcal{F}_\text{occ}$ and upsampled using $\mathcal{U}_\theta$. All the aforementioned convolutions have a filter size of 3, stride of 1 and no max pooling. Because all the tensors in this recurrence are spatial, this design choice of using interpolation and transposed convolutions to perform the hidden computations at a lower spatial resolution is important to keep the GPU memory requirements low.

\section{Additional Qualitative Results}
\begin{figure}
    \centering     %
    \subfigure[t=0s]{\label{fig:a}\includegraphics[width=60mm]{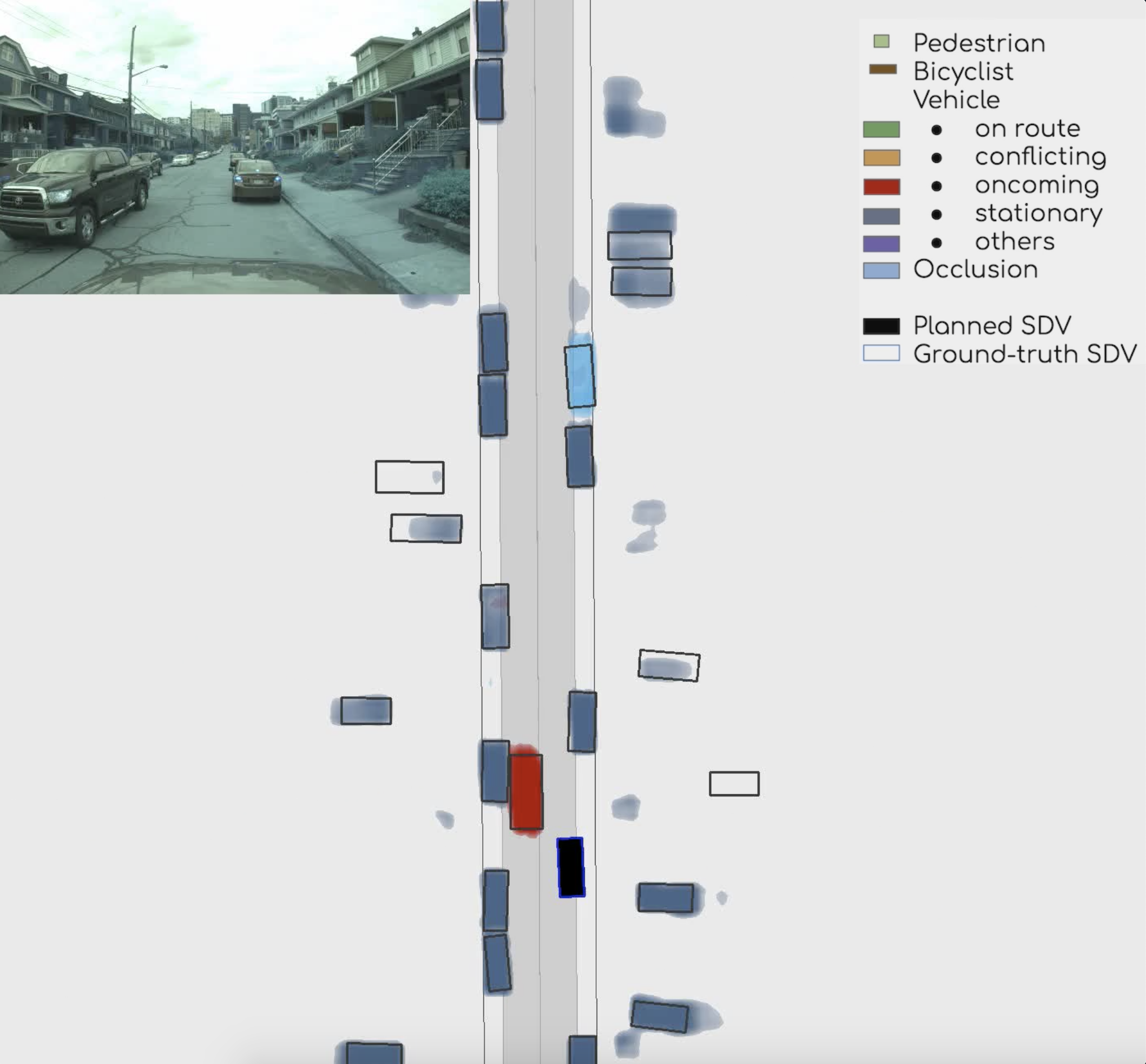}}
    \subfigure[t=1.5s]{\label{fig:b}\includegraphics[width=60mm]{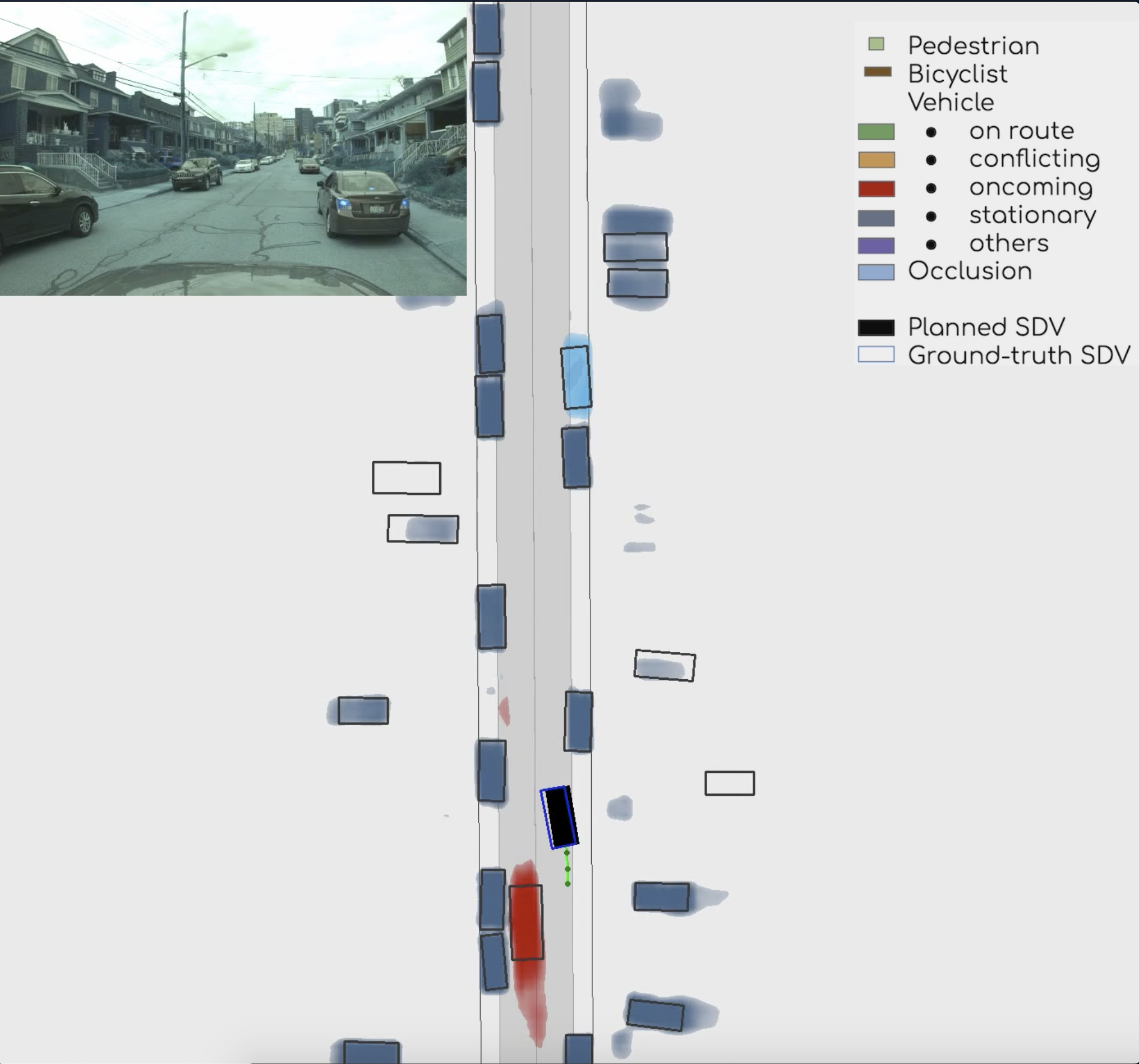}}
    \subfigure[t=3.0s]{\label{fig:c}\includegraphics[width=60mm]{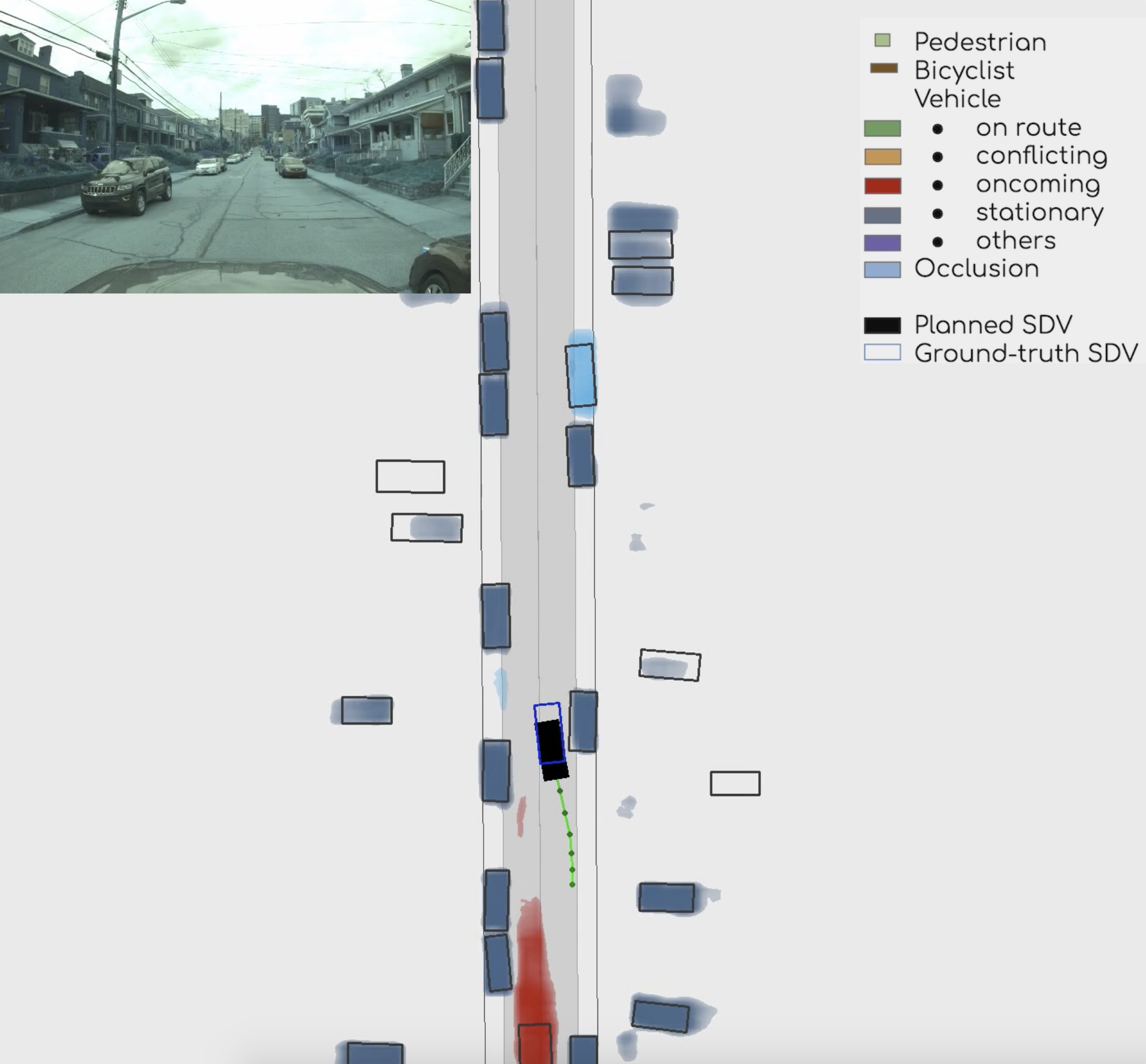}}
    \subfigure[t=4.5s]{\label{fig:d}\includegraphics[width=60mm]{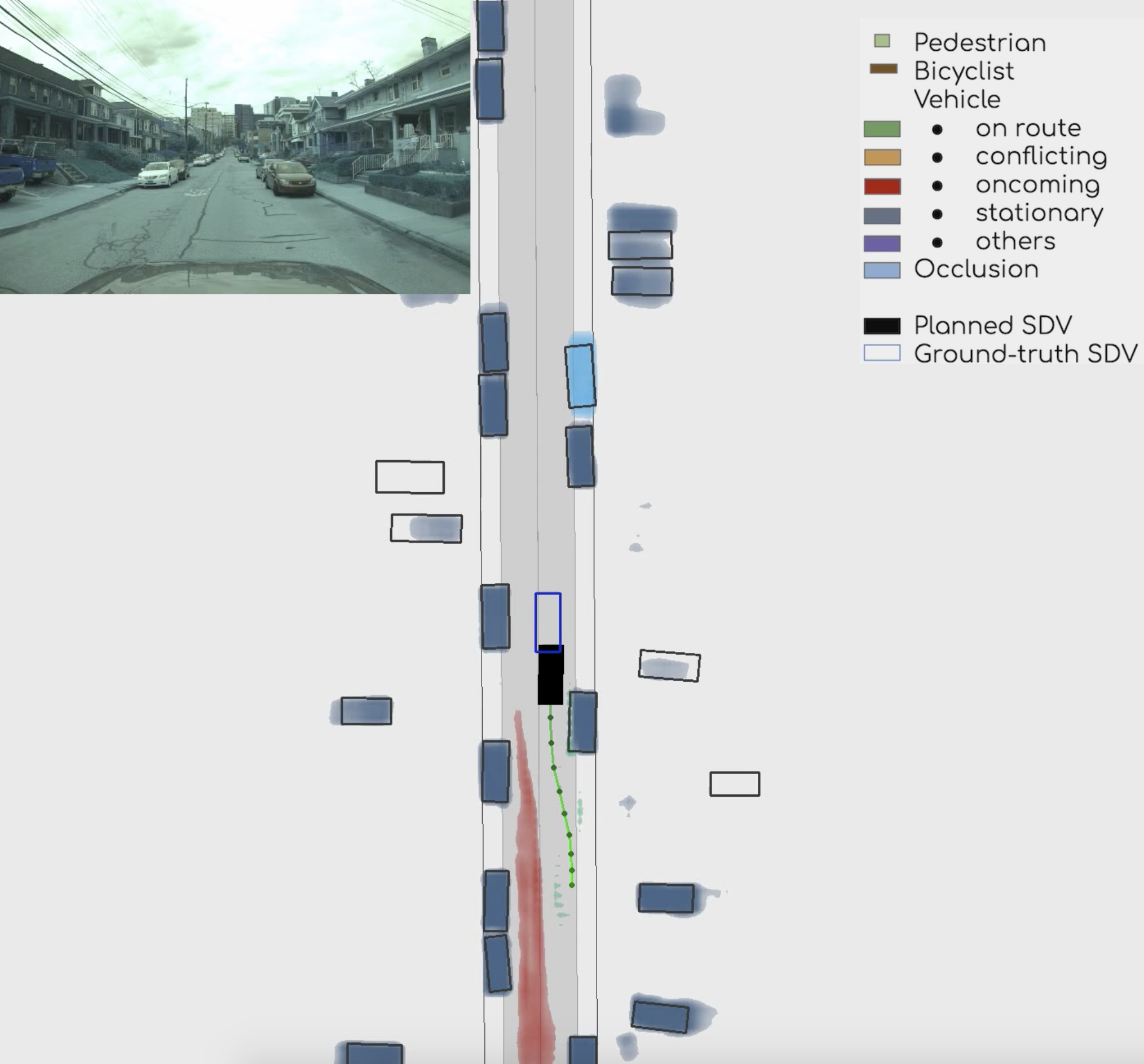}}
    \caption{\textbf{Qualitative results}: This figure demonstrates a scenario with a n oncoming truck as well as many stationary vehicles. The SDV is able to nudge around the parked vehicle adn continue in the route.}
    \label{fig:qualitative1}
\end{figure}

\begin{figure}
    \centering     %
    \subfigure[t=0s]{\label{fig:a}\includegraphics[width=60mm]{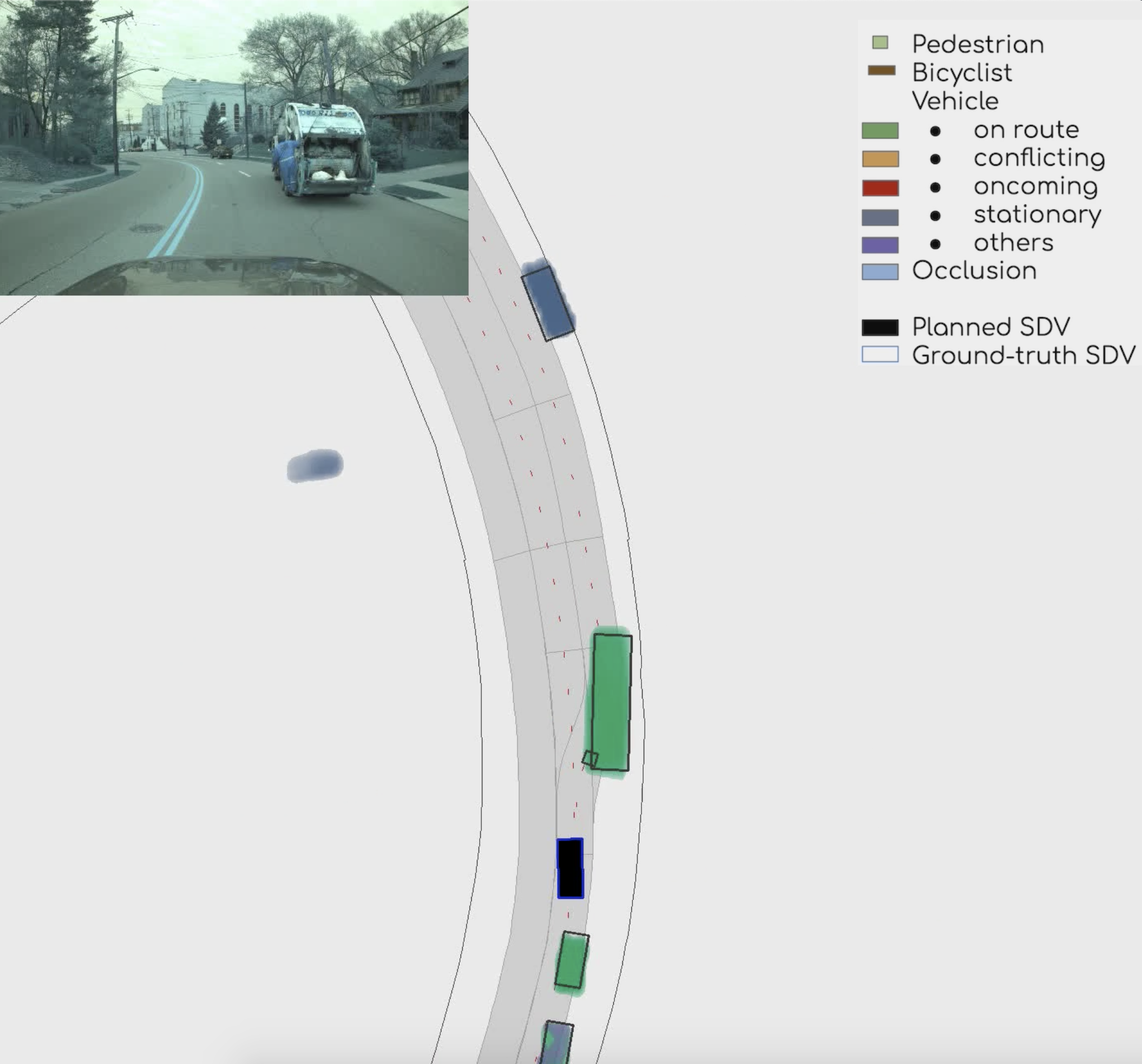}}
    \subfigure[t=1.5s]{\label{fig:b}\includegraphics[width=60mm]{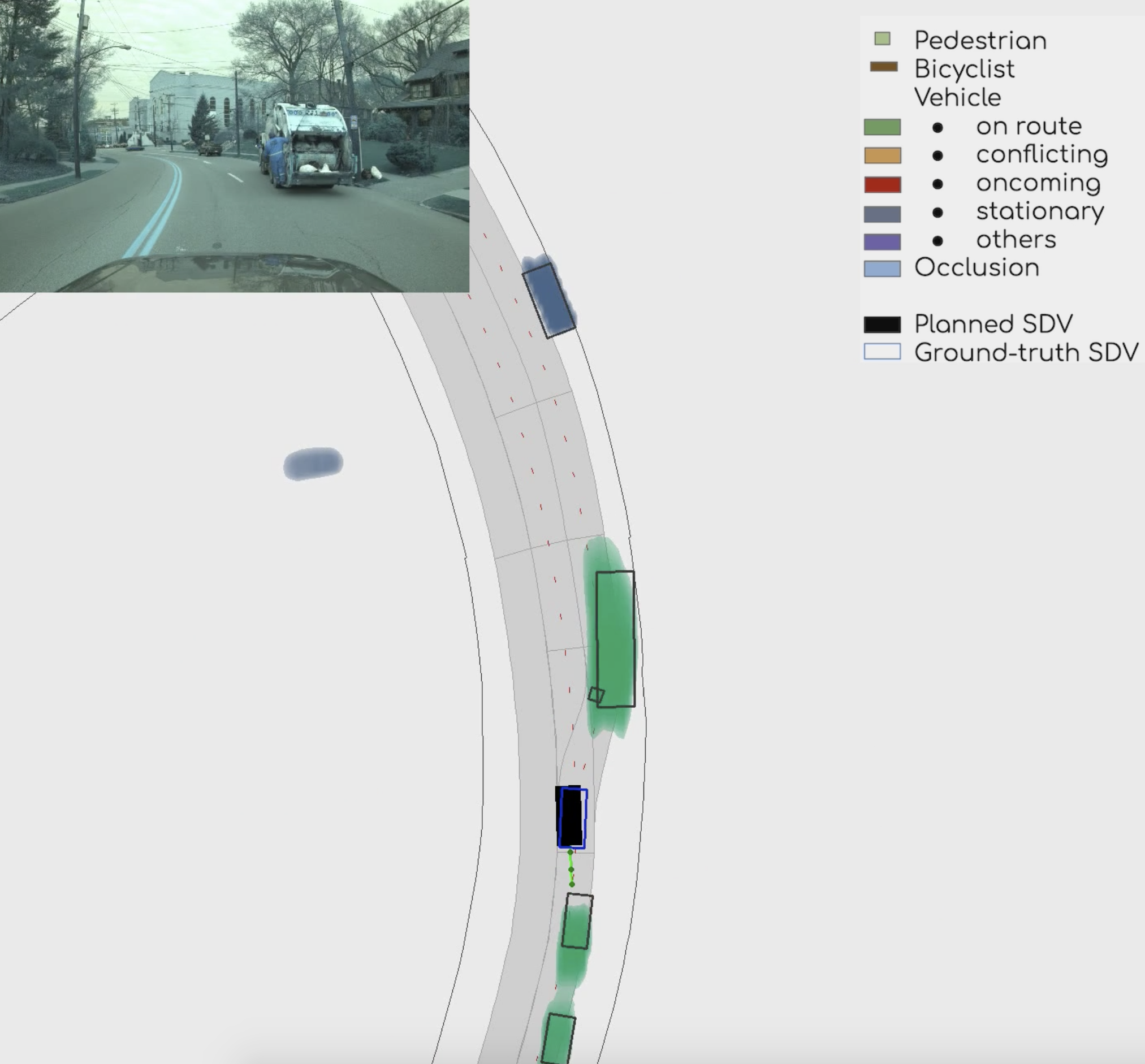}}
    \subfigure[t=3.0s]{\label{fig:c}\includegraphics[width=60mm]{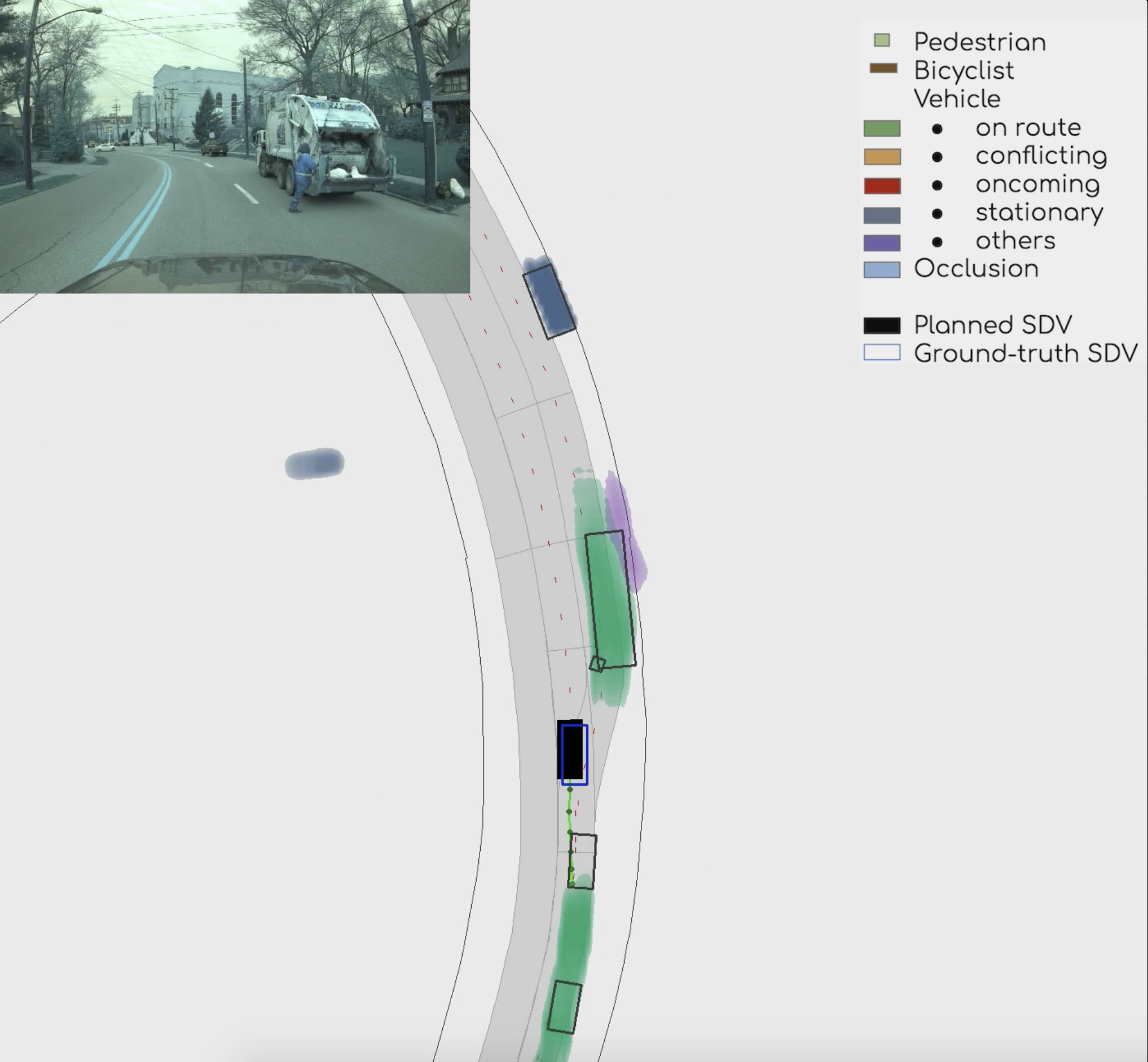}}
    \subfigure[t=4.5s]{\label{fig:d}\includegraphics[width=60mm]{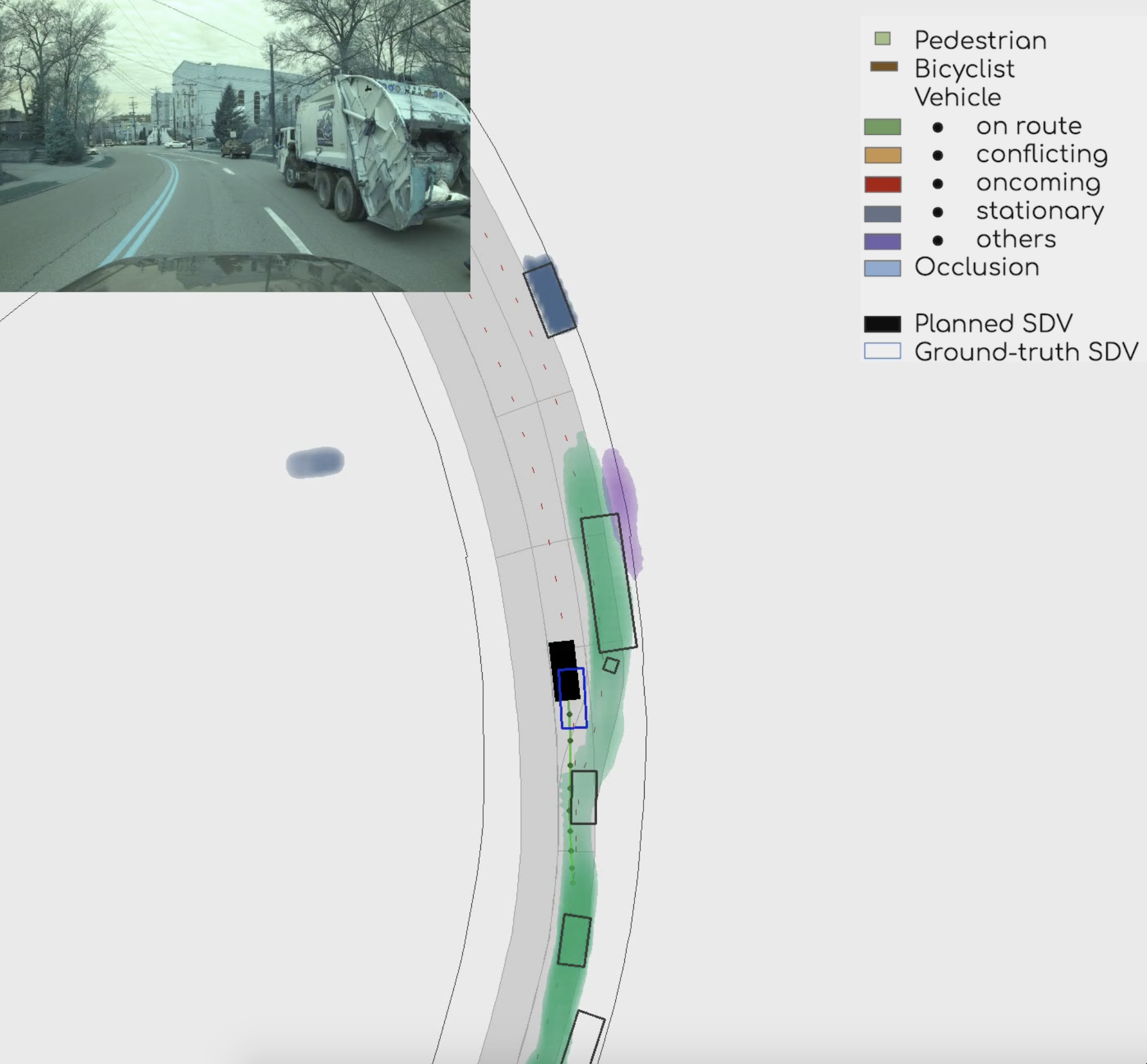}}
    \caption{\textbf{Qualitative results}: This example shows a garbage truck. As the truck is entering the right lane, and the SDV is able to nudge around it and continue in the route. Note that the occupancy representation of the truck is covering the person that on the side of the truck too.}
    \label{fig:qualitative2}
\end{figure}

\begin{figure}
    \centering     %
    \subfigure[t=0s]{\label{fig:a}\includegraphics[width=60mm]{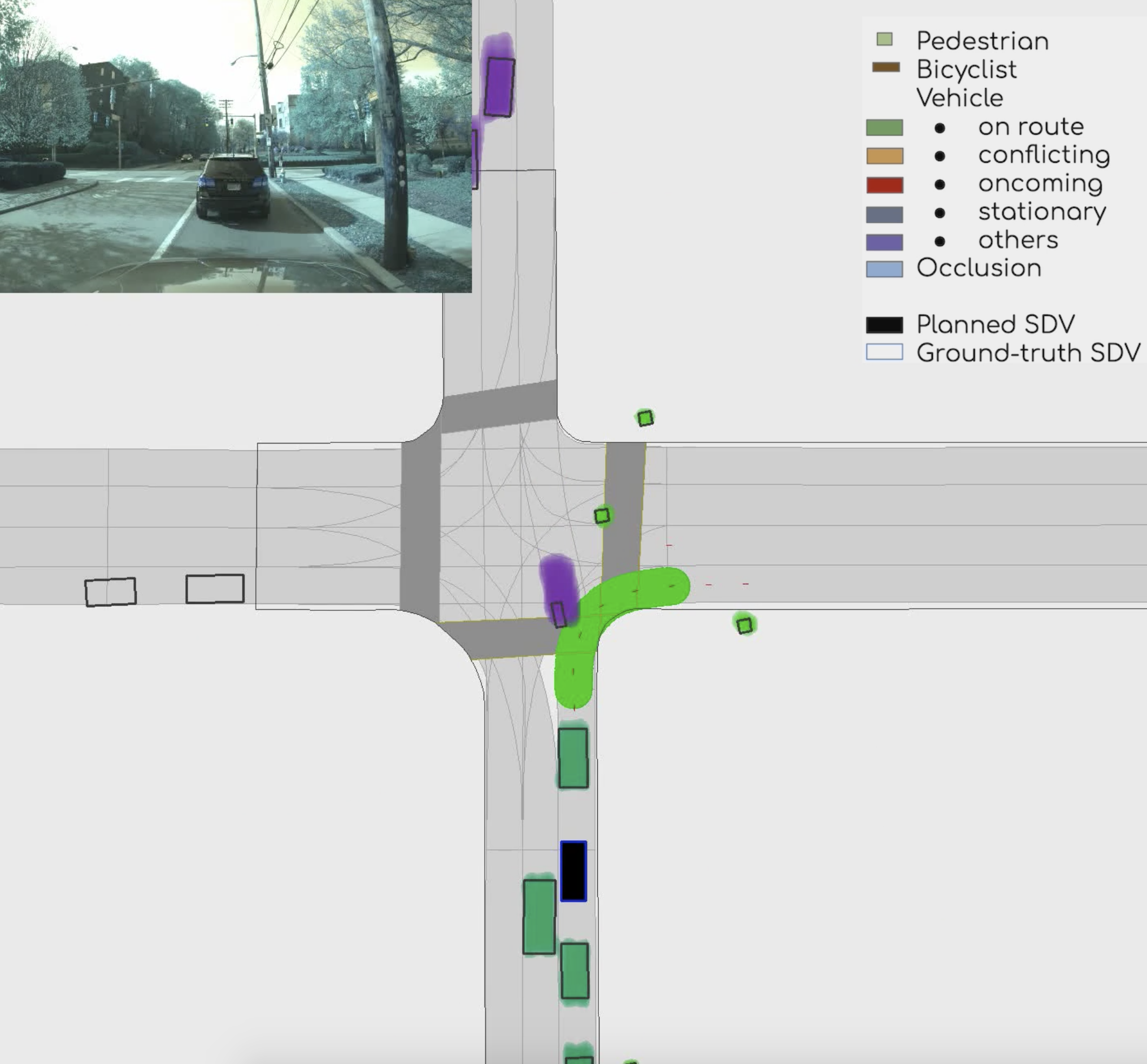}}
    \subfigure[t=1.5s]{\label{fig:b}\includegraphics[width=60mm]{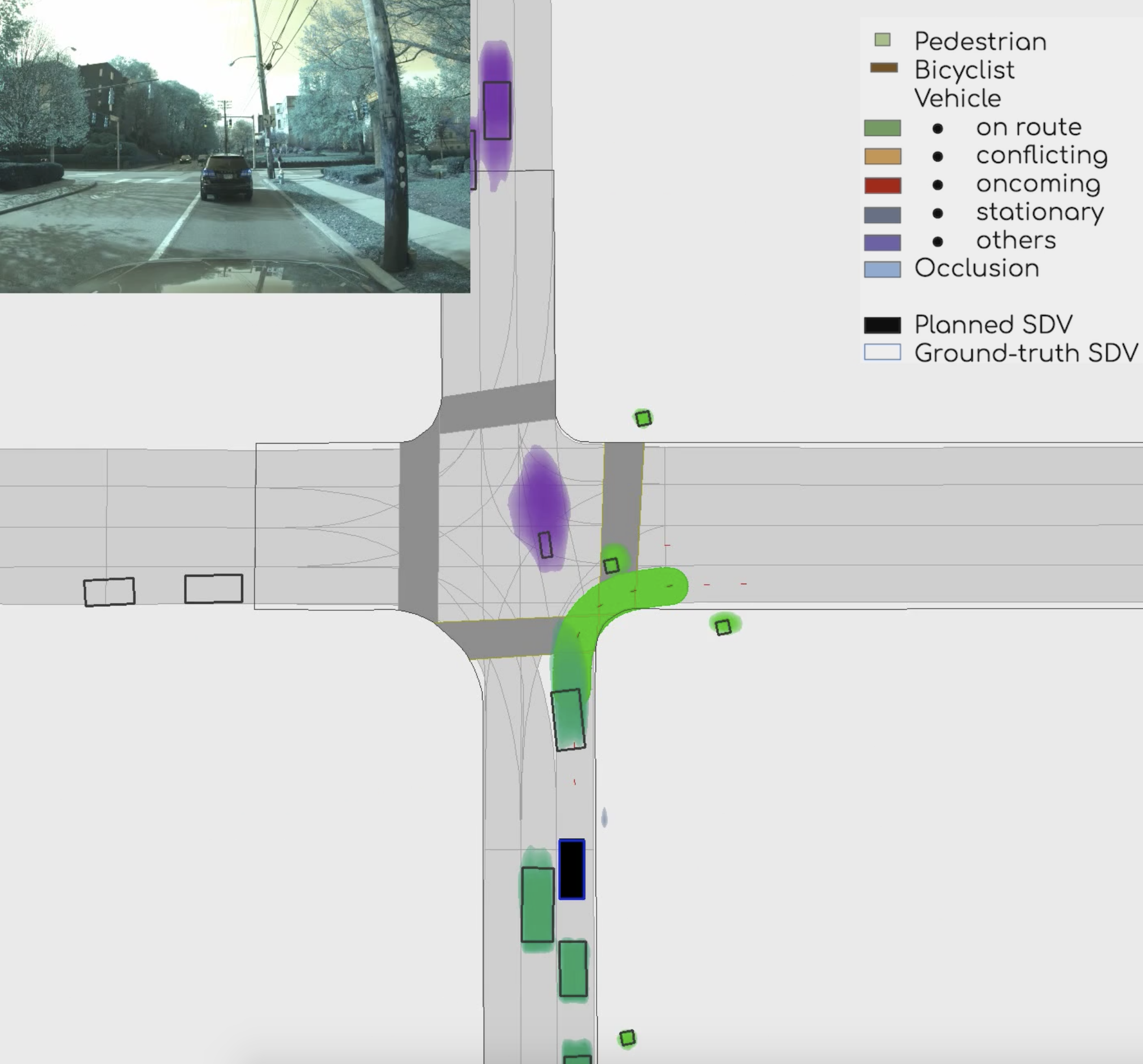}}
    \subfigure[t=3.0s]{\label{fig:c}\includegraphics[width=60mm]{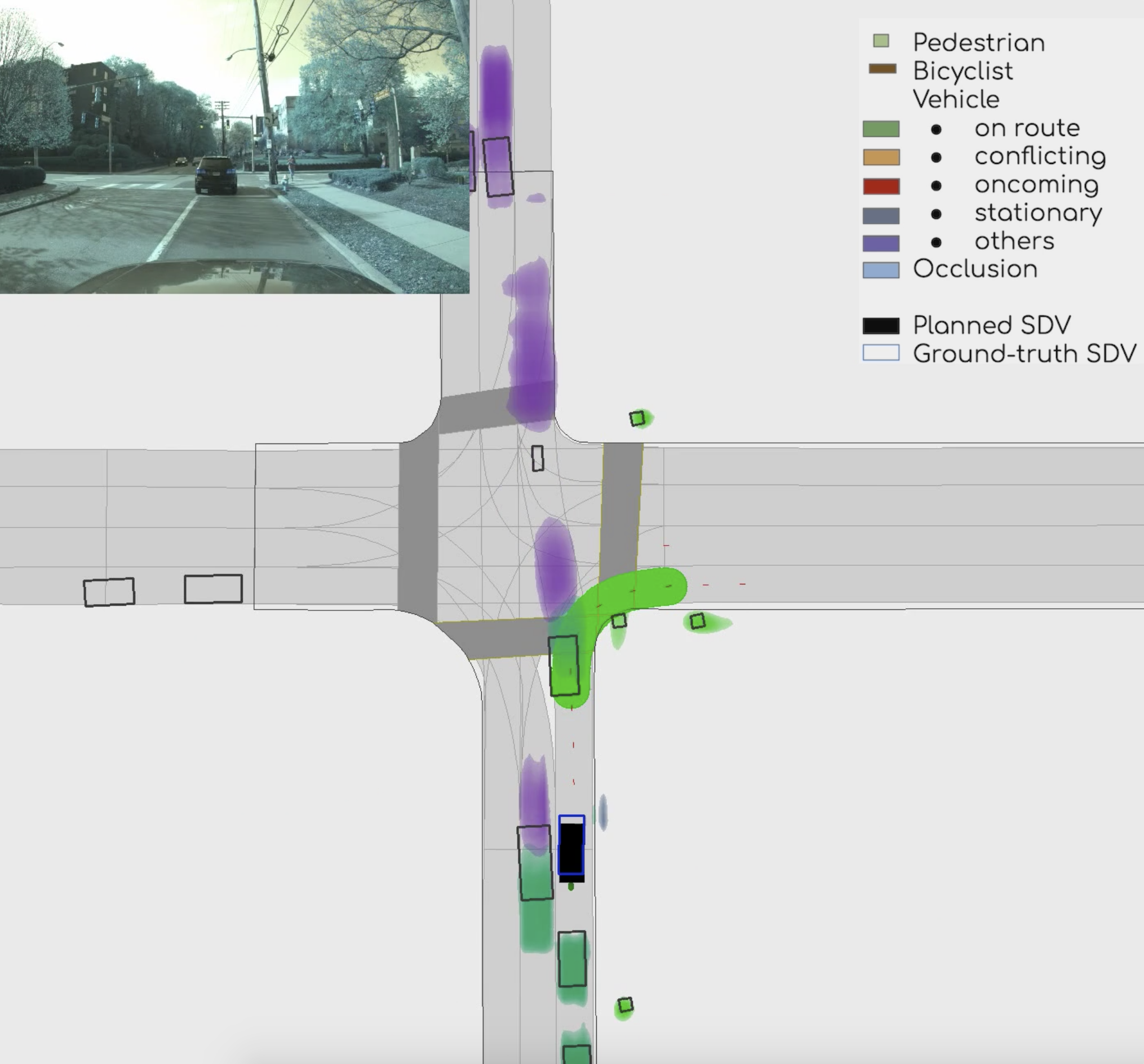}}
    \subfigure[t=4.5s]{\label{fig:d}\includegraphics[width=60mm]{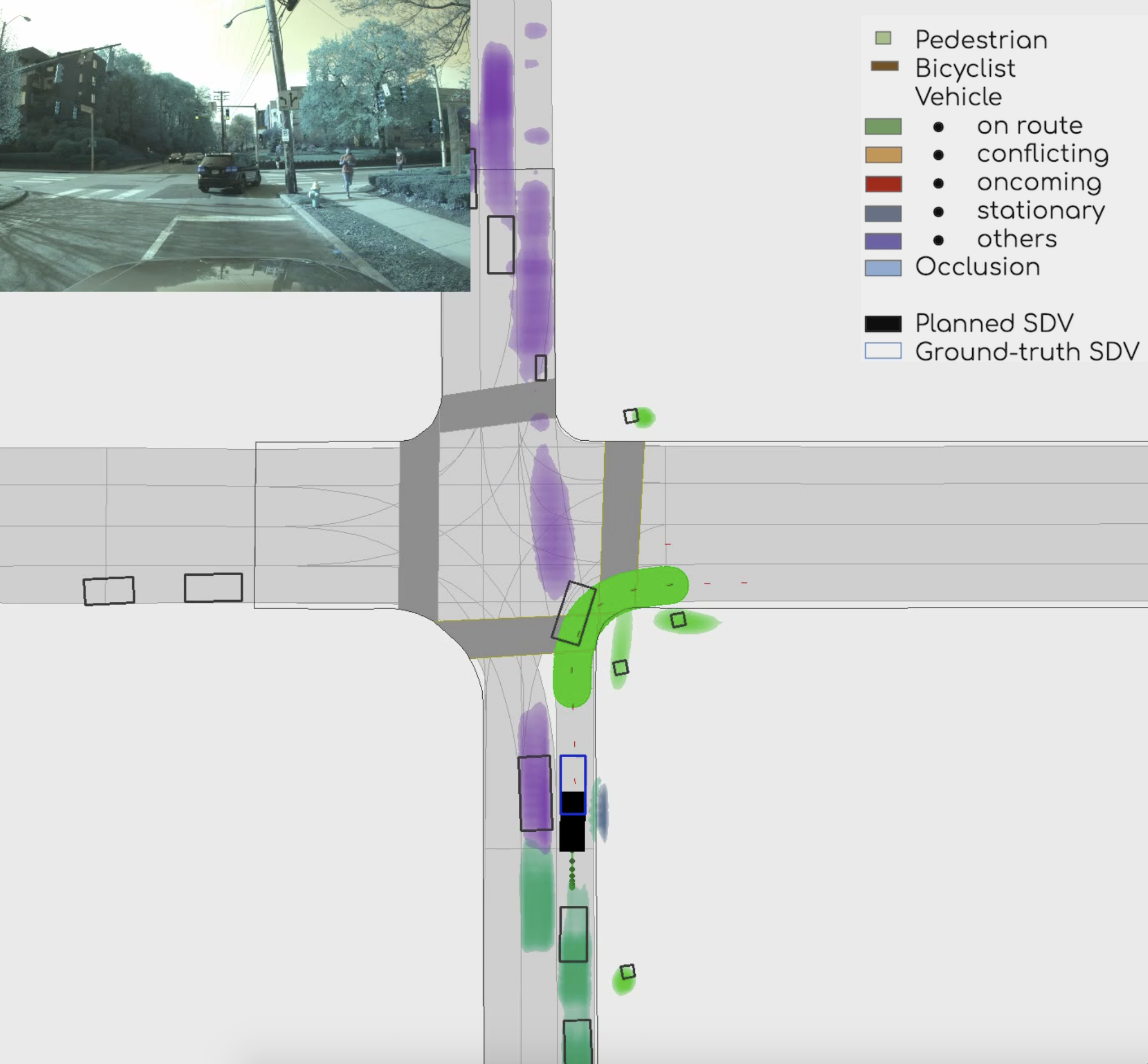}}
    \caption{\textbf{Qualitative results}: This figure show a vehicle on the left of the SDV at t=0s. As the vehicle approaches the intersection, it is categorized as others (i.e.~not relevant to the planner) as it is located on a left-turn lane.}
    \label{fig:qualitative3}
\end{figure}

\begin{figure}
    \centering     %
    \subfigure[t=0s]{\label{fig:a}\includegraphics[width=60mm]{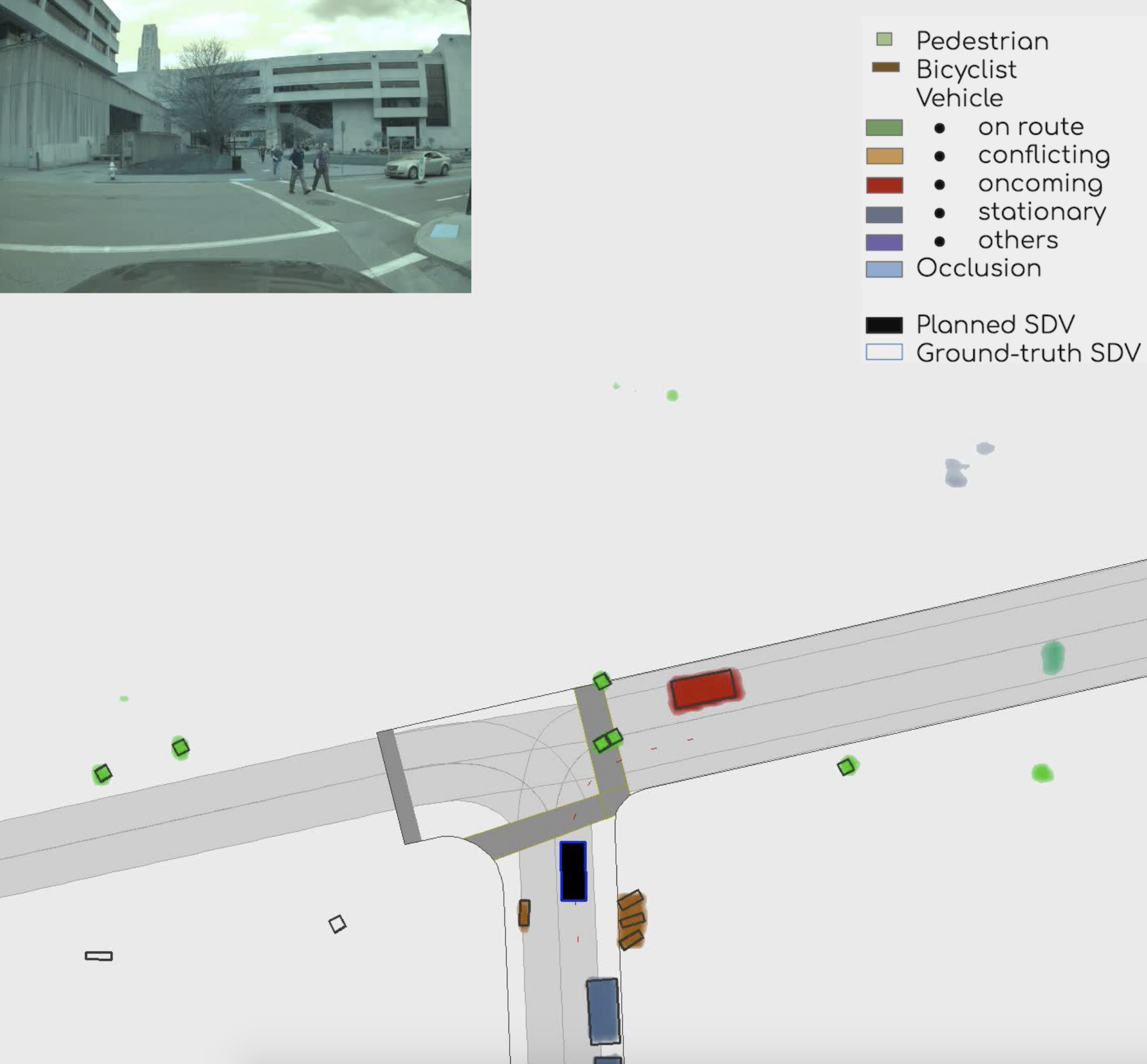}}
    \subfigure[t=1.5s]{\label{fig:b}\includegraphics[width=60mm]{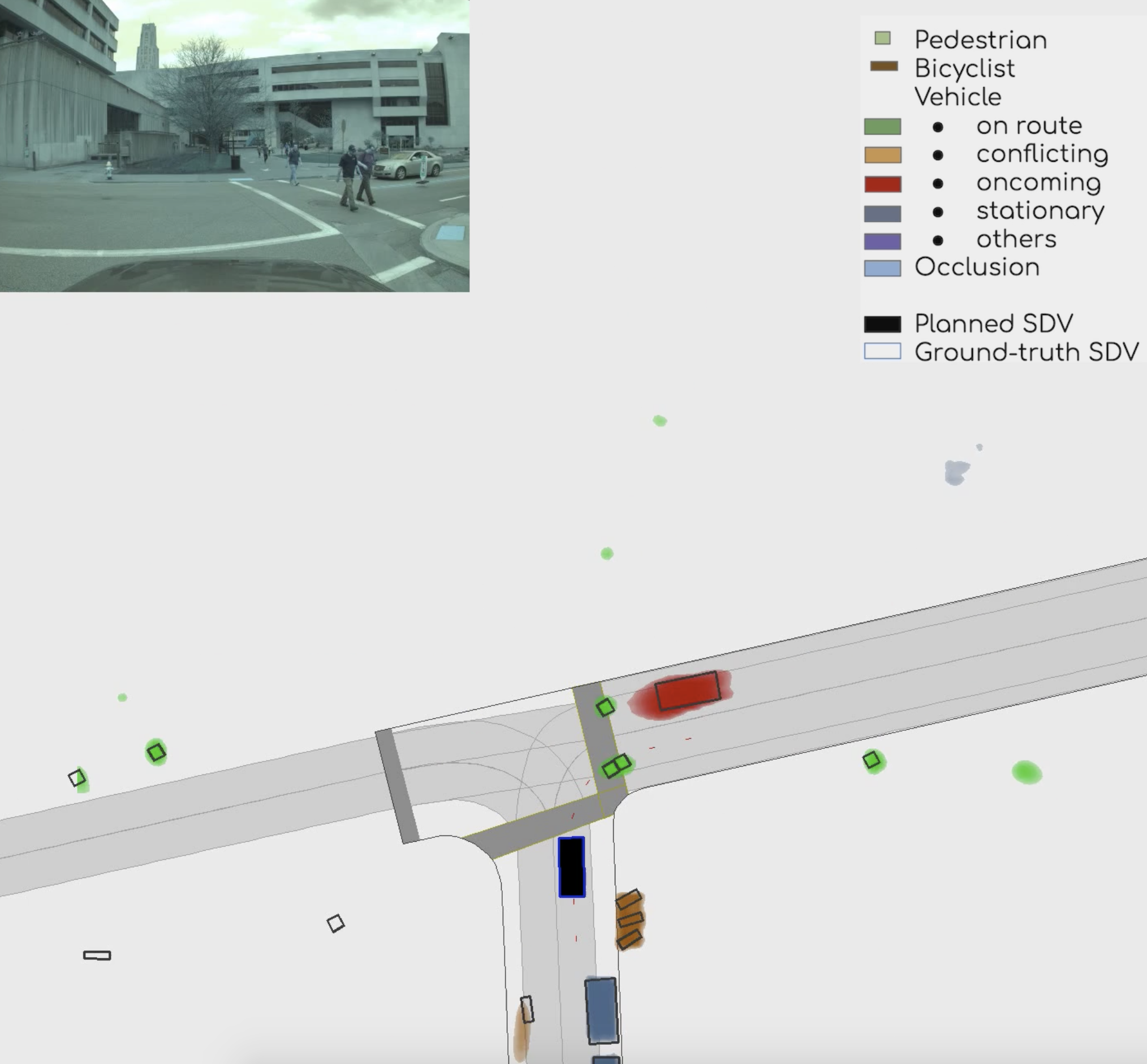}}
    \subfigure[t=3.0s]{\label{fig:c}\includegraphics[width=60mm]{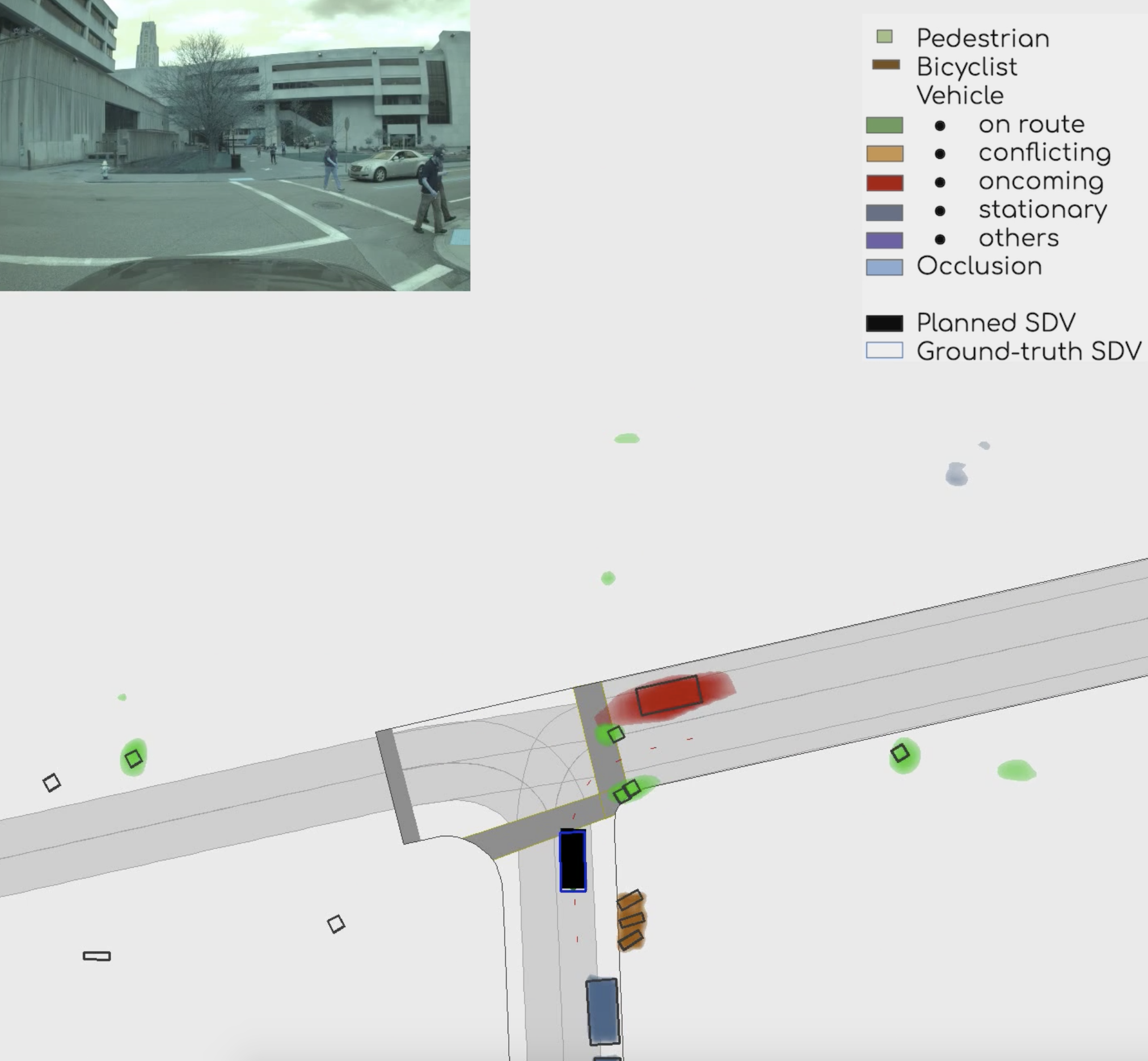}}
    \subfigure[t=4.5s]{\label{fig:d}\includegraphics[width=60mm]{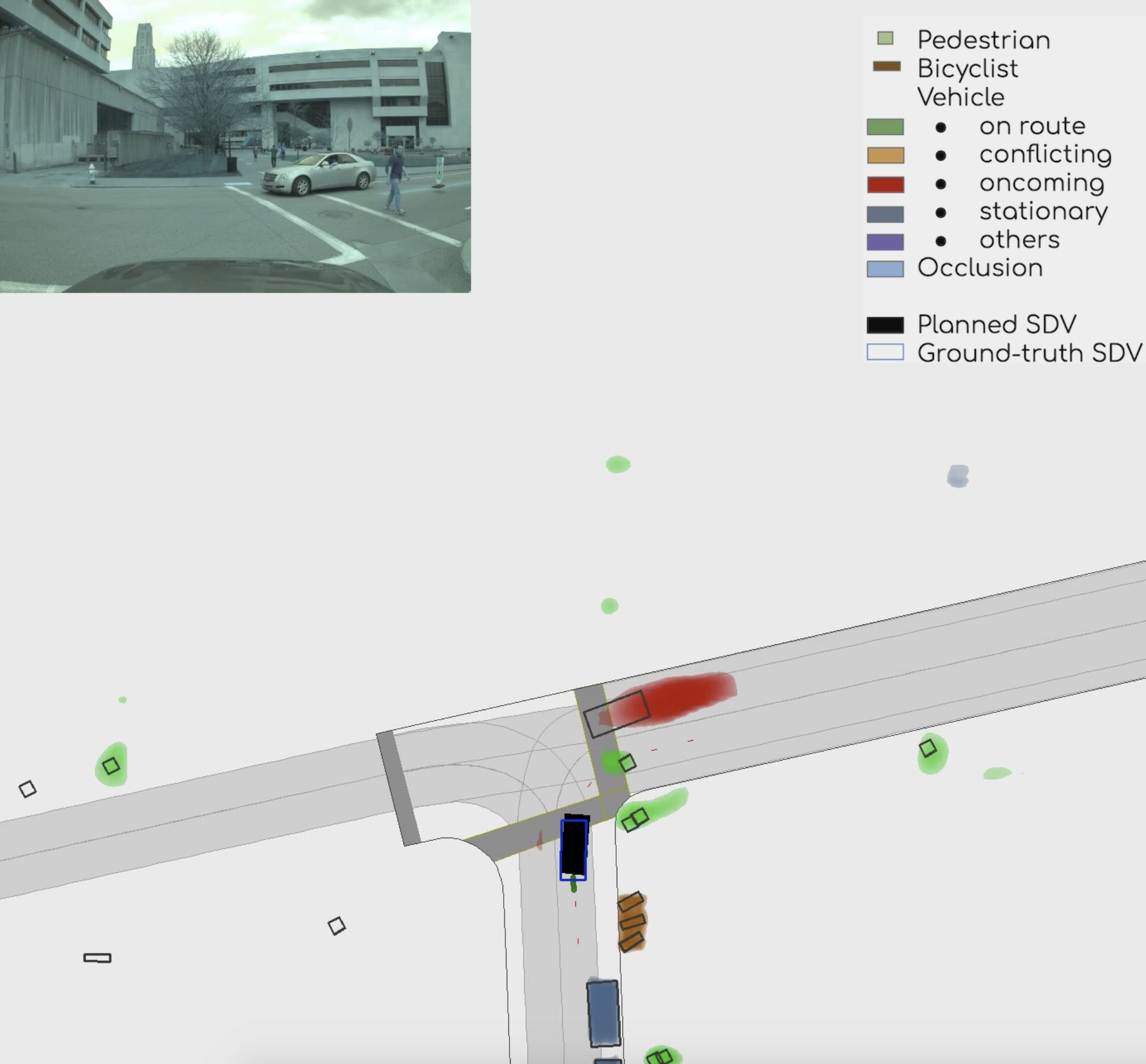}}
    \caption{\textbf{Qualitative results}: This example show cautious behavior od the SDV as some pedestrians are crossing the street.}
    \label{fig:qualitative5}
\end{figure}

\begin{figure}
    \centering     %
    \subfigure[t=0s]{\label{fig:a}\includegraphics[width=60mm]{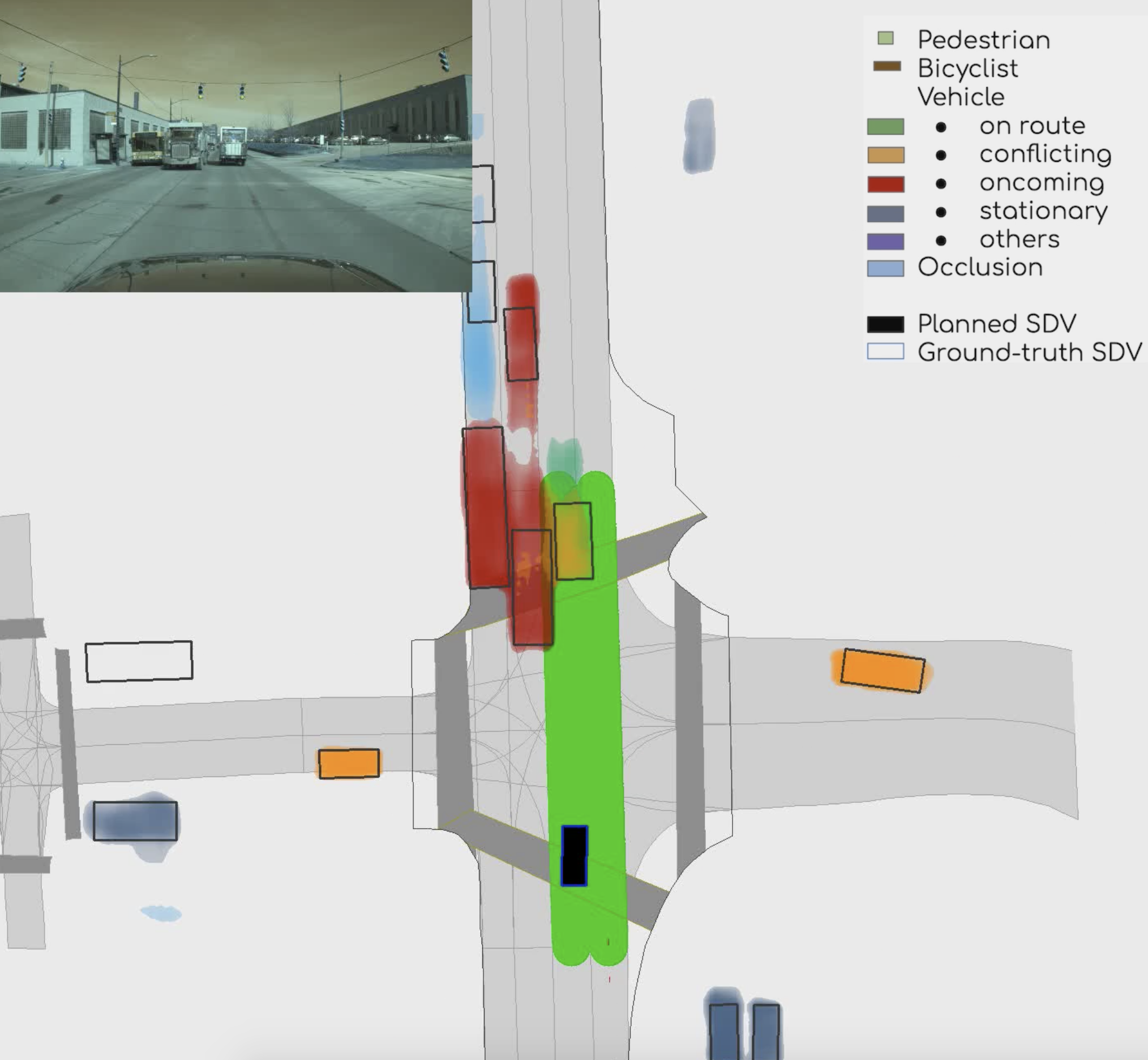}}
    \subfigure[t=1.5s]{\label{fig:b}\includegraphics[width=60mm]{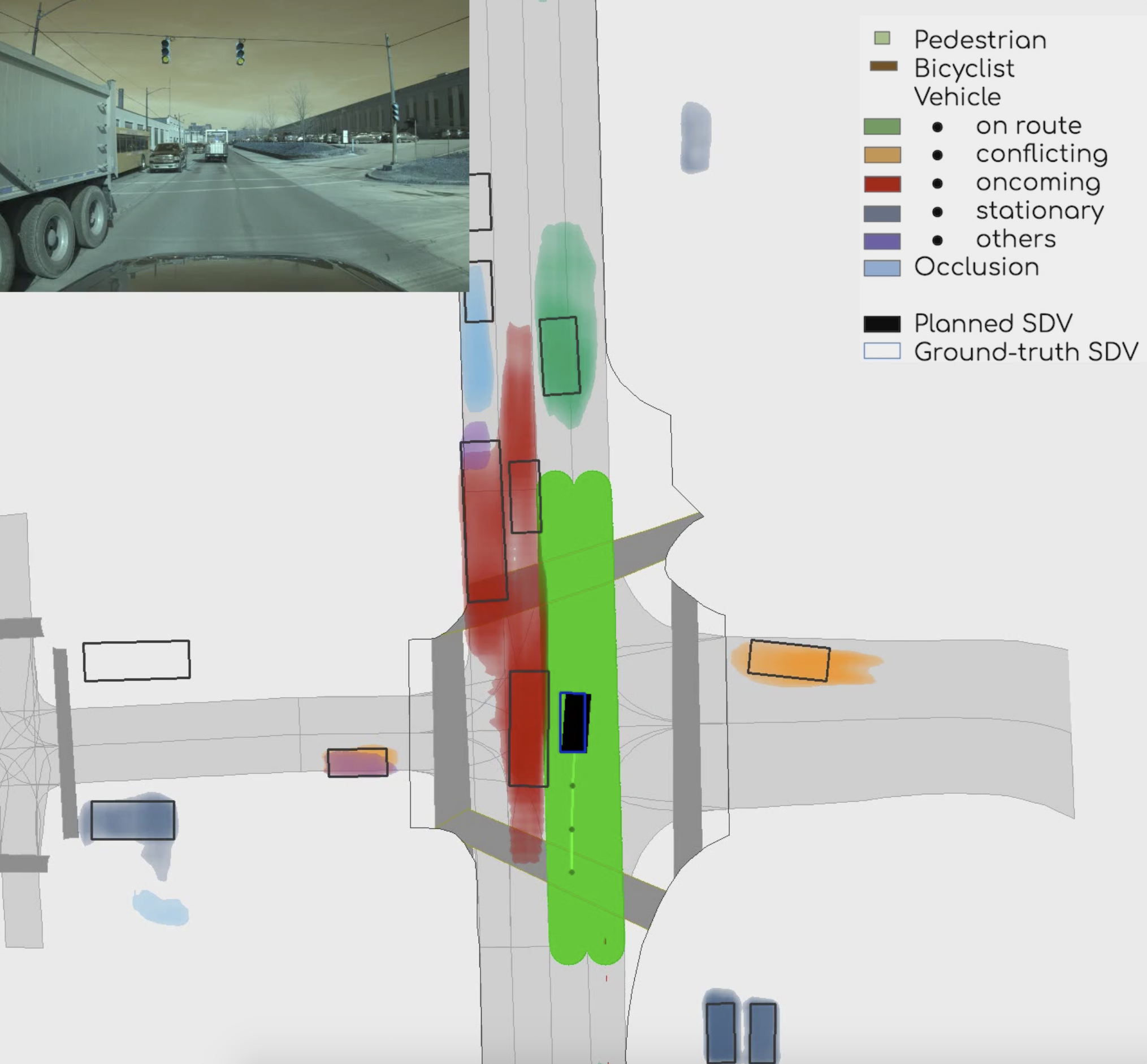}}
    \subfigure[t=3.0s]{\label{fig:c}\includegraphics[width=60mm]{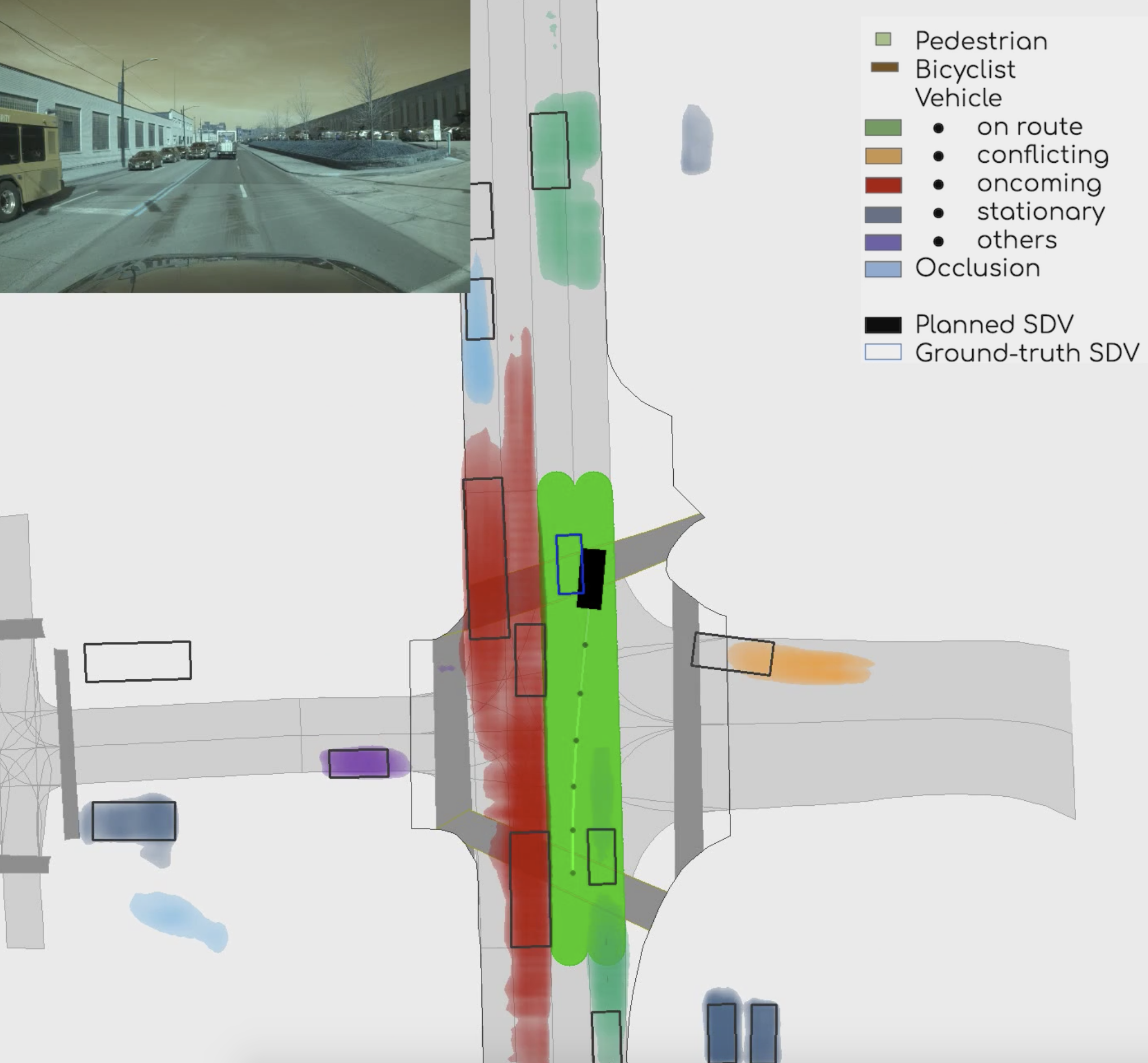}}
    \subfigure[t=4.5s]{\label{fig:d}\includegraphics[width=60mm]{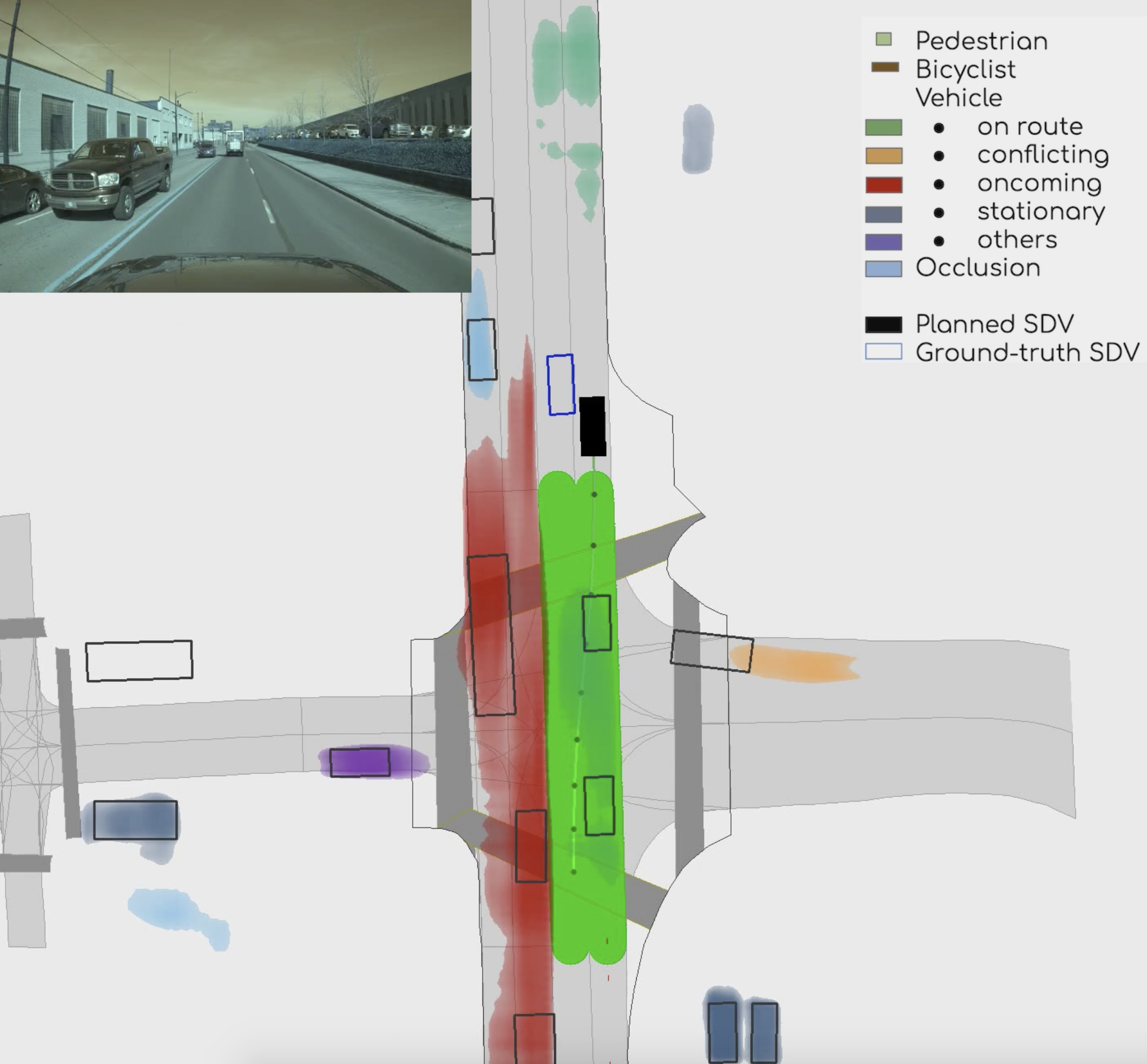}}
    \caption{\textbf{Qualitative results}: In this scenario large vehicles are occupying the oncoming lane, and a truck is encroaching the SDV lane a little bit. The planner chooses to lane-change to the right, contrary to the human driver that continued driving on the same lane.}
    \label{fig:qualitative6}
\end{figure}

Figures \ref{fig:qualitative1}-\ref{fig:qualitative6} show additional qualitative results. Each figure include the occupancy at the current time as well as the forecasts for future time-steps.
\begin{figure}
    \centering     %
    \subfigure[t=0s]{\label{fig:a}\includegraphics[width=60mm]{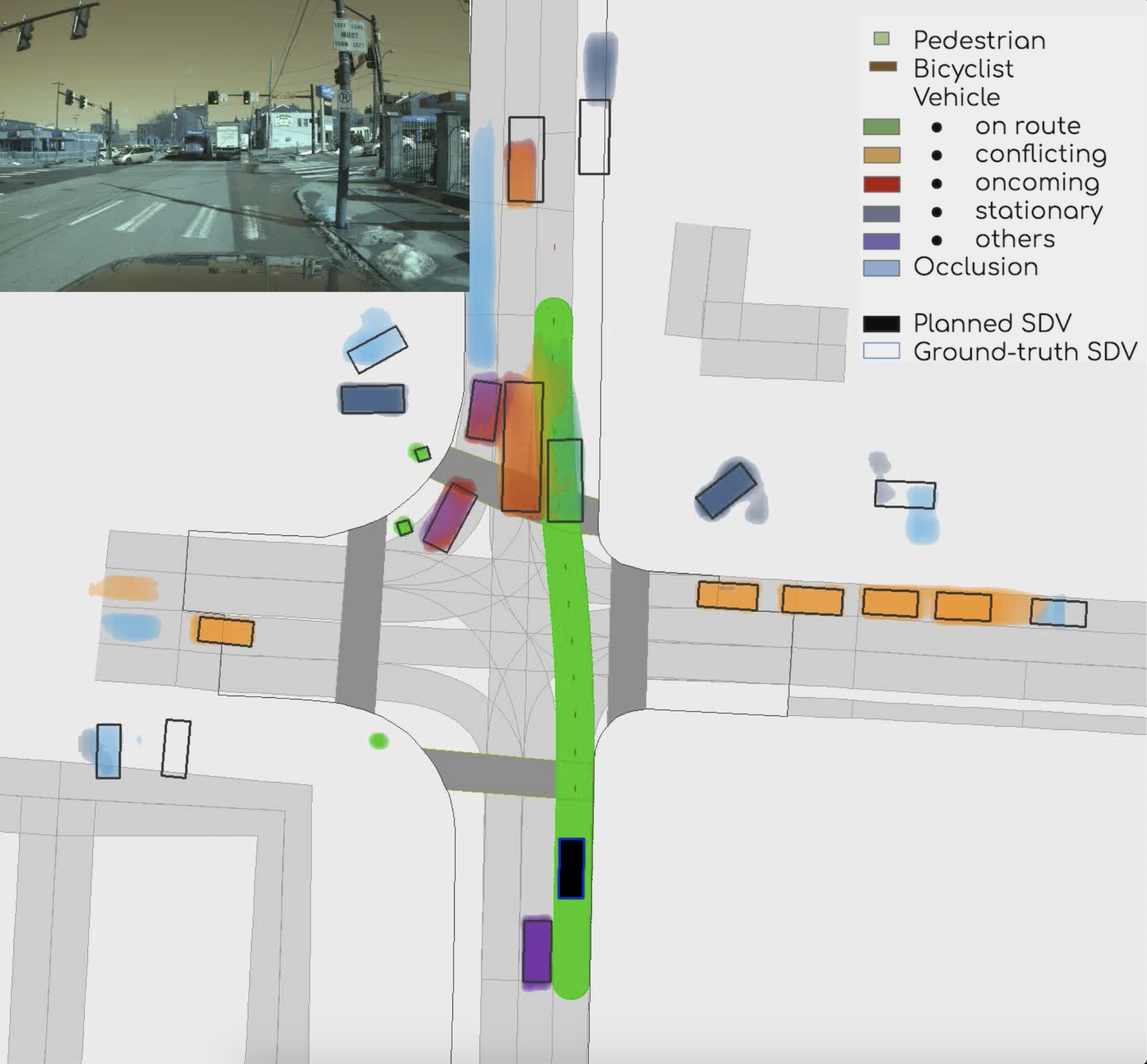}}
    \subfigure[t=1.5s]{\label{fig:b}\includegraphics[width=60mm]{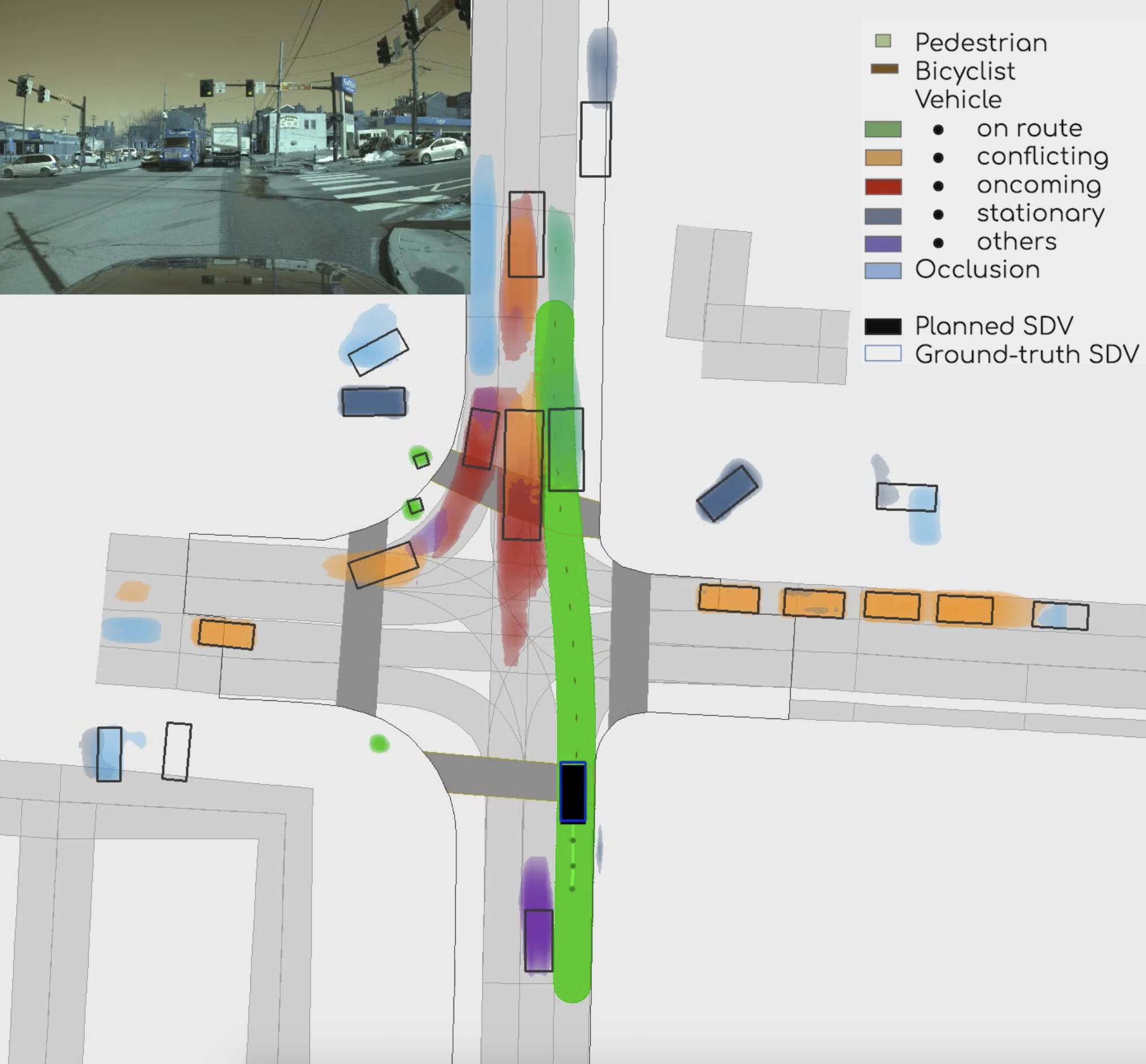}}
    \subfigure[t=3.0s]{\label{fig:c}\includegraphics[width=60mm]{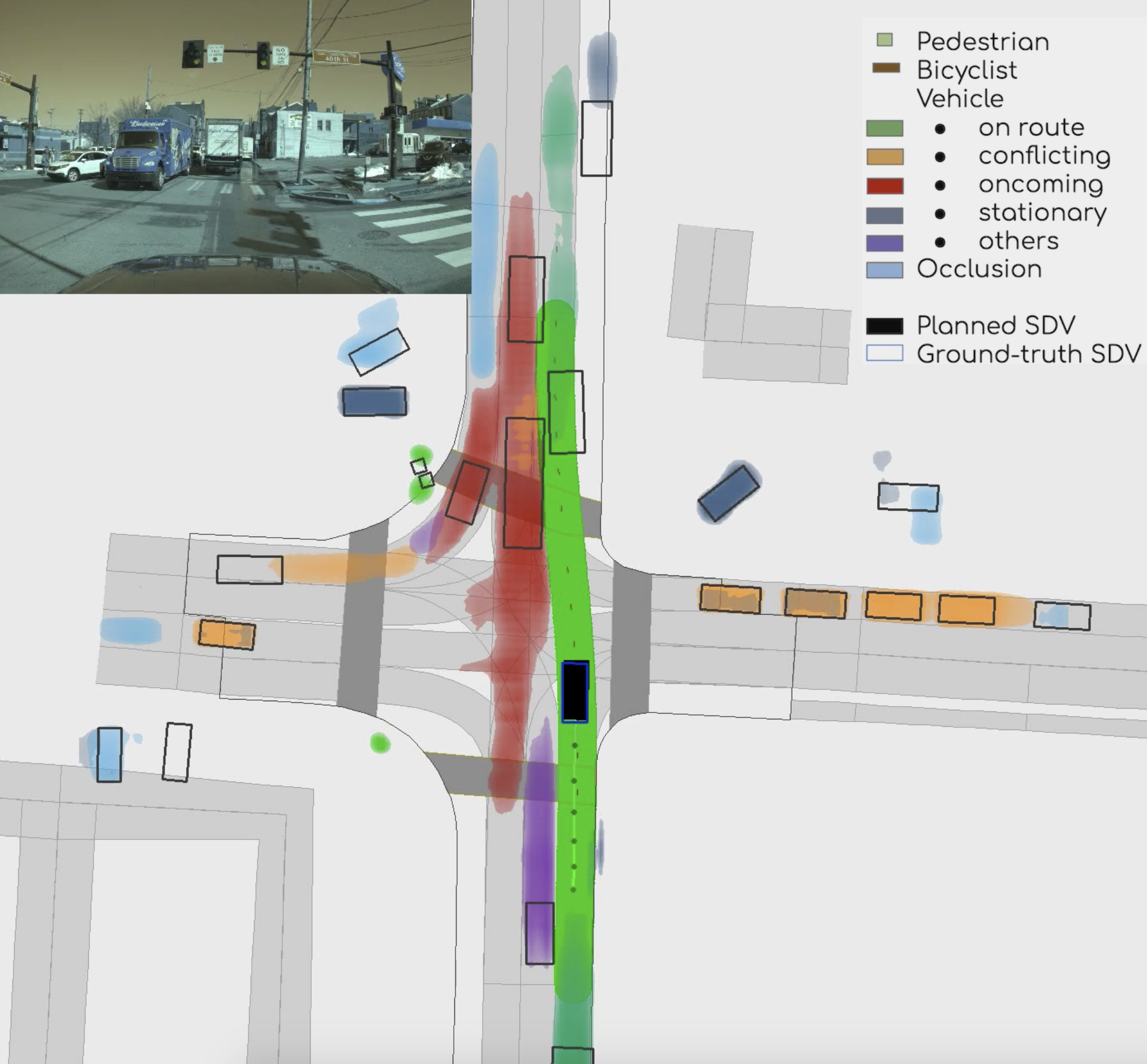}}
    \subfigure[t=4.5s]{\label{fig:d}\includegraphics[width=60mm]{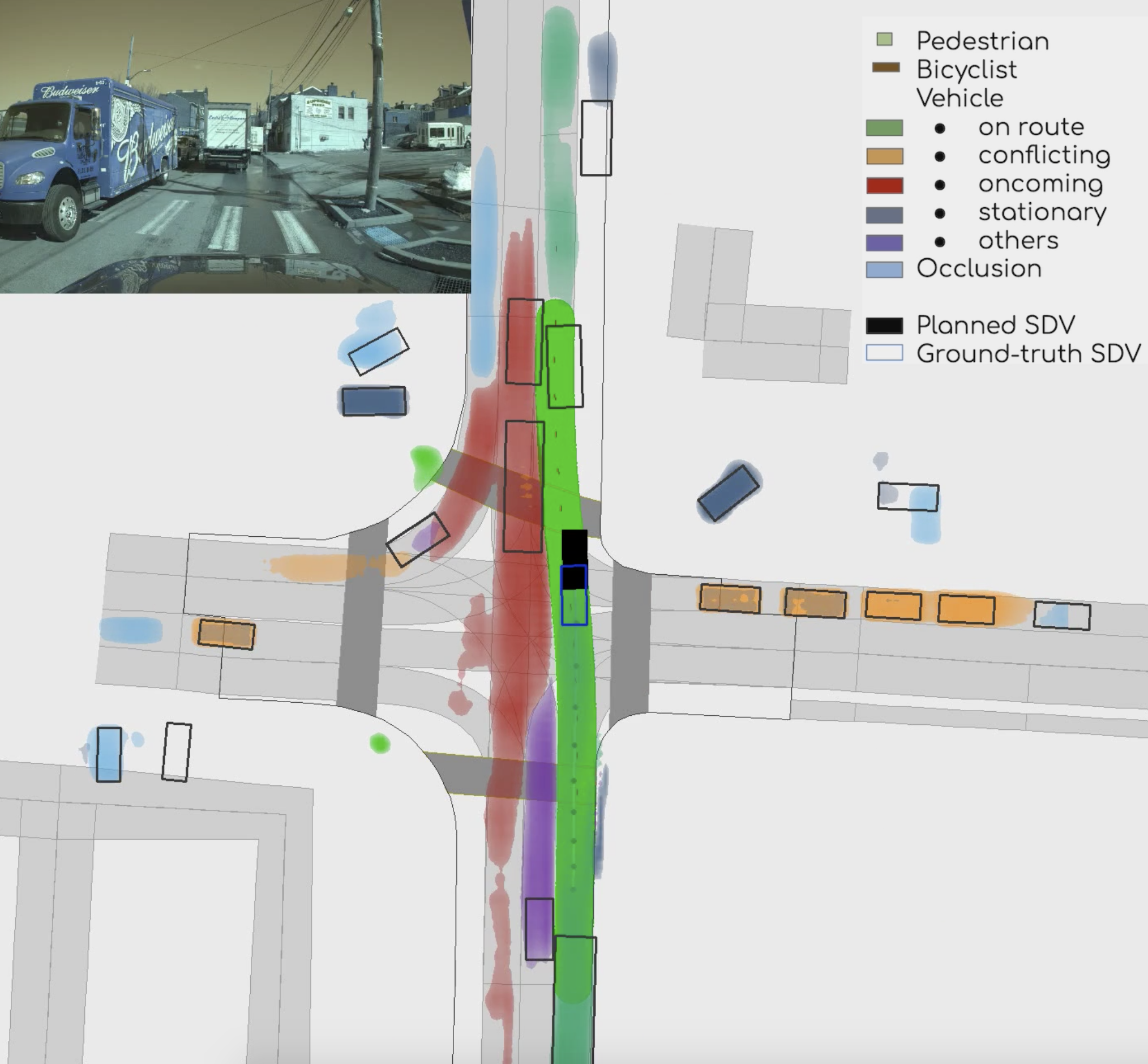}}
    \caption{\textbf{Qualitative results}: The legend on the top-right shows the color representing each subcategories of actors. On top-left, we show the image captured at the time by a front-view camera on the SDV. In this scenario, we can see regions on the lane (top-middle) that is occluded due to the obstruction by the oncoming truck.}
    \label{fig:qualitative4}
\end{figure}


\end{document}